\documentclass{tlp}

\usepackage{amssymb}
\usepackage{multirow}
\usepackage{mathtools}
\usepackage{pifont}
\newcommand{\cmark}{\ding{51}}%
\newcommand{\xmark}{\ding{55}}%
\usepackage{listings}
\usepackage{tikz}
\usetikzlibrary{shapes,decorations,arrows,calc,arrows.meta,fit,positioning,automata}
\usepackage{pgfplots}
\usepackage[colorlinks,
    linkcolor={blue},
    citecolor={blue}]{hyperref}
\usepackage{amsmath}
\usepackage{algorithm}
\usepackage{array}
\usepackage{collcell}
\usepackage{bigstrut}
\newcolumntype{I}{>{\collectcell\mathit}l<{\endcollectcell}}
\usepackage[noend]{algpseudocode}
\algrenewcommand\algorithmicensure{\textbf{Output:}}
\newcommand{\bg}{\mathit{burglary}}
\newcommand{\eq}{\mathit{earthquake}}
\newcommand{\al}{\mathit{alarm}}
\newcommand{\df}{\mathit{defective}}
\newcommand{\rt}{\mathit{right}}
\newcommand{\hx}{\mathit{stay\_at\_X}}
\newcommand{\hy}{\mathit{stay\_at\_Y}}
\newcommand{\te}{\mathit{too\_expensive\_X}}
\newcommand{\tn}{\mathit{too\_noisy\_Y}}

\usepackage{caption}
\usepackage{subcaption}

\usepackage{xcolor}
\newcommand{\smp}[0]{\textsc{sm}ProbLog}
\newtheorem{example}{Example}
\newtheorem{definition}{Definition}

\newenvironment{program}
    {\begin{center}
    \begin{tabular}{c}
    \hline\\[-1em]$
    }
    { 
        $
    \\\hline
    \end{tabular} 
    \end{center}
    }

\begin{document}

\lefttitle{P.Totis et. al.}

\jnlPage{1}{8}
\jnlDoiYr{2021}
\doival{10.1017/xxxxx}

\title[\smp{}]
{\smp{}: Stable Model Semantics in ProbLog for Probabilistic Argumentation}

\begin{authgrp}
    \author{\sn{Pietro} \gn{Totis} }
    \affiliation{KU Leuven, Dept. of Computer Science; Leuven.AI, B-3000 Leuven, Belgium\\
        \email{pietro.totis@cs.kuleuven.be}}
    \author{\sn{Luc} \gn{De Raedt}\footnote{\label{note}Shared last authors}}
    \affiliation{KU Leuven, Dept. of Computer Science; Leuven.AI, B-3000 Leuven, Belgium\\
        \email{luc.deraedt@cs.kuleuven.be}}
    \author{\sn{Angelika} \gn{Kimmig}\textsuperscript{\ref{note}}}%

    \affiliation{KU Leuven, Dept. of Computer Science; Leuven.AI, B-3000 Leuven, Belgium\\
        \email{angelika.kimmig@cs.kuleuven.be}}

\end{authgrp}

\history{\sub{23 November 2021;} \rev{xx xx xxxx;} \acc{xx xx xxxx}}

\maketitle

\begin{abstract}
    Argumentation problems are concerned with determining the acceptability of a set of arguments from their relational structure.
    When the available information is uncertain, probabilistic argumentation frameworks provide modelling tools to account for it.
    The first contribution of this paper is a novel interpretation of probabilistic argumentation frameworks as probabilistic logic programs.
    Probabilistic logic programs are logic programs in which some of the facts are annotated with probabilities.
    We show that the programs representing probabilistic argumentation frameworks do not satisfy a common assumption in probabilistic logic programming (PLP) semantics, which is, that probabilistic facts fully capture the uncertainty in the domain under investigation.
    The second contribution of this paper is then a novel PLP semantics for programs where a choice of probabilistic facts does not uniquely determine the truth assignment of the logical atoms.
    The third contribution of this paper is the implementation of a PLP system supporting this semantics: \smp{}.
    \smp{} is a novel PLP framework based on the probabilistic logic programming language ProbLog.
    \smp{} supports many inference and learning tasks typical of PLP, which, together with our first contribution, provide novel reasoning tools for probabilistic argumentation.
    We evaluate our approach with experiments analyzing the computational cost of the proposed algorithms and their application to a dataset of argumentation problems.

    \emph{Under consideration in Theory and Practice of Logic Programming (TPLP).}
\end{abstract}


\section{Introduction}
Designing systems that are able to argue and persuade is a very relevant and challenging task in artificial intelligence.
In real-world scenarios the information available is often incomplete or questionable, therefore the ability to take into account uncertainty is fundamental in such situations.
Probabilistic logic programming (PLP) frameworks and languages, such as PRISM~\citep{DBLP:conf/ilp/SatoK08}, ICL~\citep{DBLP:conf/ilp/Poole08}, ProbLog~\citep{DBLP:conf/ijcai/RaedtKT07} or LPAD/CP-logic~\citep{DBLP:journals/tplp/VennekensDB09}, are designed to provide powerful general-purpose tools for modelling and reasoning about structured, uncertain domains.
On the other hand, abstract argumentation frameworks~\citep{DBLP:journals/ai/Dung95} aim specifically at describing argumentation processes and reasoning about the set of acceptable arguments.
Consider, for instance, the following example of argumentative microtext adapted from~\citep{DBLP:conf/lrec/StedeAPAP16}:
\begin{example}\label{ex:microtext}
    Yes, it's annoying and cumbersome to separate your rubbish properly all the time ($a_1$), but small gestures become natural by daily repetition ($a_2$). Three different bin bags stink away in the kitchen and have to be sorted into different wheelie bins ($a_3$). But still Germany produces way too much rubbish ($a_4$) and too many resources are lost when what actually should be separated and recycled is burnt ($a_5$). We Berliners should take the chance and become pioneers in waste separation! ($a_6$)
\end{example}
An argumentation problem defines arguments, e.g. $a_1,a_2,\dots$ and their relations, for example $a_1$ attacks $a_2$,  $a_3$ supports $a_1$\dots from which a set of acceptable arguments is derived according to a given semantics.

When uncertainty is part of the argumentation problem, we want to reason about it in order to answer queries like ``What is the likelihood of accepting argument $a_6$?''. Inferring the probability of one variable taking a particular value is a typical PLP task. PLP frameworks offer a wide range of tools and algorithms for probabilistic inference and learning on probabilistic logic programs.
Probabilistic argumentation systems, on the other hand, propose different combinations of argumentation frameworks, probability interpretations and reasoning systems, tailored to manipulating probabilities in argumentation.
When considering PLP and argumentation systems, the question is natural: can PLP effectively model argumentation processes and reason over its intrinsic uncertainty?


This question has been only partially investigated so far: of the two main interpretations of probabilities in argument graphs~\citep{DBLP:journals/ijar/Hunter13}, namely the \emph{constellations} approach and the \emph{epistemic} approach, only the former has been studied in the context of PLP~\citep{DBLP:journals/ijar/MantadelisB20}. This paper fills this gap and shows how PLP can be used to reason about a probabilistic argumentation graph with an epistemic view of its probabilities. Table~\ref{tab:compare_args_overview} summarizes the differences in the approaches.

\begin{table}[b]
    \centering
    \caption{Argumentation frameworks overview.}
    \label{tab:compare_args_overview}
    \begin{tabular}{ c  c  c  c}
        \topline
        Framework                                            & Epistemic & PLP    & EM Learning \midline
        MetaProblog~\citep{DBLP:journals/ijar/MantadelisB20} & \xmark    & \cmark & \xmark               \\
        Epistemic graphs~\citep{DBLP:journals/ai/HunterPT20} & \cmark    & \xmark & \xmark               \\
        \smp{}                                               & \cmark    & \cmark & \cmark
        \botline
    \end{tabular}
\end{table}

In the constellations approach, probabilities describe a probability distribution over argument (sub)graphs. Such a distribution is used to define the probability of an argument being in the set of accepted arguments according to the argumentation semantics of choice.
On the contrary, in the epistemic approach probabilities are a direct measure of the belief of an agent in an argument, and therefore of its acceptability. Previous work on probabilistic argument graphs under the epistemic approach focuses on modelling a family of probability distributions that are compatible with the graph and given constraints~\citep{DBLP:journals/ai/HunterPT20}.

Instead of considering a probabilistic argument graph as a model for a family of probability distributions over fixed beliefs that have to satisfy constraints, we regard a probabilistic argument graph as a model for a single joint probability distribution over arguments.
We view nodes as random variables associated to the arguments, whose prior is the epistemic arbitrary assignment of probabilities to arguments.
We then interpret their relations as the conditional (in)dependencies influencing the posterior joint probability (belief) distribution of the random variables. Given
a set of independent beliefs regarding arguments and relations, modelled by means of PLP, we thus define the joint probability (belief) of the arguments by means of probabilistic inference.

The joint distribution defines \emph{marginal and conditional} beliefs that are coherent with the given priors and argumentative (logic) structure.
The advantage of this method is that the joint belief distribution allows us to answer a broad range of questions that are of interest in the argumentation setting. For example: the marginal probability of (belief in) an argument (``What is the belief in $a_1$ considering the influence of the other arguments?''), relating the beliefs of two arguments (``How does my belief in $a_1$ change if $a_5$ is accepted?''), or perform typical PLP tasks such as learning the beliefs of an agent given a set of observations of accepted arguments.

We propose a new PLP framework where this is possible, since we argue that existing PLP systems are limited in what problems they can model and reason about.
In fact, traditional PLP frameworks do not allow modelling cyclic relations involving negations. This is a pattern often found in argument graphs, where reciprocal attacks are common. For example, we will encode this pattern, i.e. accepting $a_1$ inhibits my belief in $a_2$ and vice versa, with the logic rules $\lnot a_1\leftarrow a_2.\; \lnot a_2 \leftarrow a_1$. A model containing such rules would not be valid in traditional PLP frameworks, because they assume that each (deterministic) logic program induced by the probabilistic model has exactly one two-valued well-founded model~\citep{DBLP:journals/jacm/GelderRS91}. Programs containing cyclic dependencies involving negation usually do not satisfy this assumption. Probabilistic Answer Set Programming frameworks, e.g.~\citeN{DBLP:journals/tplp/BaralGR09} or \citeN{DBLP:conf/kr/LeeW16}, do not make this assumption, however the definition of the probability distribution is not based on distribution semantics~\citep{DBLP:conf/iclp/Sato95}, hence there are differences that we will analyze throughout the paper. Table~\ref{tab:compare_PLP_overview} summarizes the main characteristics of PLP frameworks.

\begin{table}[t]
    \caption{PLP frameworks overview.}
    \label{tab:compare_PLP_overview}
    \centering
    \begin{tabular}{ c  p{3cm}  c}
        \topline
        Framework                                          & Model semantics                      & Distribution semantics \midline
        PLP frameworks (ProbLog, PRISM, \dots)             & Least or 2-valued Well-founded model & \cmark                          \\
        Probabilistic ASP (P-Log, LP\textsuperscript{MLN}) & Stable model                         & \xmark                          \\
        \smp{}                                             & Stable model                         & \cmark
        \botline
    \end{tabular}
\end{table}

In Section~\ref{sec:args} we introduce a framework modelling probabilistic argumentation problems by means of typical PLP modelling techniques. In Section~\ref{sec:joint} we analyze the properties of this approach. In Sections~\ref{sec:semantics} and~\ref{sec:implementation} we propose a PLP system, \smp{}, based on a new semantics, where it is possible to reason over such models.
The advantages of reasoning over probabilistic argument graphs by means of PLP are:
\begin{itemize}
    \item \emph{Succintness and expressivity}: expressing complex interactions between random variables with the simplicity of (first-order) logic programming rules.
    \item \emph{Flexibility and modularity}: improving a model for a specific application domain without having to modify neither the language nor the reasoning algorithms.
    \item \emph{PLP tools}: applying to argumentation problems a general suite of PLP inference and learning algorithms to any model within a broad class.
\end{itemize}

To summarize, the key contributions of the paper are:
\begin{enumerate}
    \item We define a novel semantics for PLP based on stable model semantics~\citep{DBLP:conf/iclp/GelfondL88} (Section~\ref{sec:semantics}).
    \item  We develop an implementation of a PLP framework, \smp{}\footnote[2]{https://github.com/PietroTotis/smProblog}, derived from ProbLog2 \citep{DBLP:conf/pkdd/DriesKMRBVR15} that supports inference and learning tasks under the new semantics (Section~\ref{sec:implementation}).
    \item  We show an application of \smp{} to encode and solve probabilistic argumentation problems in a novel reasoning framework based on an epistemic interpretation of probability (Sections~\ref{sec:args} and~\ref{sec:joint}).
\end{enumerate}

\section{Background}\label{sec:background}
In this paper we present a PLP framework, \smp{} based on ProbLog2: we describe the relevant background about syntax, the semantics for the probabilistic component of ProbLog programs and the semantics for the models of a logic program we will consider throughout the paper. We also describe the corresponding inference and learning tasks under the new semantics. Finally, we also consider the application of \smp{} to probabilistic argumentation problems, hence we discuss the required background in argumentation.

\subsection{Probabilistic Logic Programming}


\paragraph{ProbLog.} ProbLog~\citep{DBLP:conf/ijcai/RaedtKT07} is a probabilistic language extending Prolog, where facts and clauses are annotated with (mutually independent) probabilities.
Probabilistic logic programs based on distribution semantics, such as ProbLog, can be viewed as a ``programming language'' generalization of Bayesian Networks~\citep{DBLP:conf/iclp/Sato95}.
A \emph{normal} logic program $\mathcal{L}$ is a finite set of normal rules of the form $ h \leftarrow b_1,\dots,b_{n}$. We use the standard terminology about atoms, terms, predicates, and literals. The \emph{head} $h$ is an atom and the \emph{body} $b_1,\dots,b_n$ is a logical conjunction of literals. Rules with empty body are called facts. We denote with ${\sim}a$ the negation as failure of an atom $a$, used to make literals, and with $\lnot a$ the classical logical negation of an atom $a$.

We model uncertainty in logic programs by annotating facts with probabilities: the probabilities of facts are mutually independent and each \emph{probabilistic fact}, i.e. $p::f.$, corresponds to an atomic choice, that is, a choice between including $f$ in the program (with probability $p$) or discarding it ($1-p$). A \emph{probabilistic normal logic program } $\mathcal{L}=F\cup R$ is thus a set $F$ of facts annotated with probabilities plus a set of logic rules $R$. The set of facts $F$ is disjoint from the heads of the rules $R$, nonetheless rule heads can be annotated as syntactic sugar: $p::h\leftarrow b_1,\dots,b_n.$ is equivalent to a new fact $p::f.$ plus $h\leftarrow f, b_1,\dots,b_n.$

\begin{example}\label{ex:alarm}
    The well-known alarm Bayesian network~\citep{DBLP:books/daglib/0066829} can be encoded as a ProbLog program.
    An alarm is triggered by either an earthquake or a burglary: if the neighbour is at home you will receive a call.
    The neighbour at home, the earthquake, and the burglary are independent events occurring with a certain probability, thus encoded as probabilistic facts. Probabilistic facts are marginally independent and therefore correspond to nodes of the Bayesian network that do not have common parents or ancestors. The logic rules determine whether you get a call or not, and thus correspond to the conditional dependencies of the Bayesian network.
    \begin{program}
        \begin{array}{l}
            0.1::burglary.                                          \\
            0.2::earthquake.                                        \\
            0.5::neighbour\_at\_home.                               \\
            alarm \leftarrow earthquake.                            \\
            alarm \leftarrow burglary.                              \\
            neighbour\_calls \leftarrow alarm, neighbour\_at\_home. \\
            query(alarm).                                           \\
            query(neighbour\_calls).
        \end{array}
    \end{program}
\end{example}

\paragraph{Distribution Semantics.} A probabilistic logic program $\mathcal{L} =F\cup R$ is queried for the likelihood of atoms. The probability of an atom is commonly defined by the \emph{distribution semantics}~\citep{DBLP:conf/iclp/Sato95}. A \emph{total choice} $\omega$ is a combination of atomic choices over all probabilistic facts, that is, a subset of the set of all ground facts $F$, i.e. $\omega\subseteq F$. We denote the set of all total choices of $\mathcal{L}$, i.e. the power set of $F$, with $\Omega_\mathcal{L}$.
The probabilities of the facts define a probability distribution over the $2^{|F|}$ total choices. The possible non-probabilistic subprograms $\omega\cup R, \omega \in\Omega_{\mathcal{L}}$, which we call \emph{possible worlds}, are obtained from $\mathcal{L}$ by including or discarding a fact according to the corresponding atomic choice. An interpretation is an assignment of a truth value to all atoms in the Herbrand base of a program. An interpretation is called a model of the theory if it satisfies all formulas in the theory.

\begin{example}\label{ex:possiblew}
    Consider the alarm example (Example~\ref{ex:alarm}), there are $2^3$=$8$ possible worlds (listed in Example~\ref{ex:alarm_pw}), corresponding to the combinations of choices for the probabilistic facts $\mathit{earthquake}$, $\mathit{burglary}$, $\mathit{neighbour\_at\_home}$. Consider the possible world where we discard $\mathit{earthquake}$ and include the other two: then the alarm will be triggered, i.e. $\mathit{alarm}$ is inferred from $\mathit{burglary}$, and thus the neighbour calls, because both $\mathit{alarm}$ and $\mathit{neighbour\_at\_home}$ are true. The model for the possible world denoted by $\{\mathit{burglary},\mathit{neighbour\_at\_home}\}$ is thus $\{\mathit{burglary}, \mathit{neighbour\_at\_home}, \mathit{alarm}, \mathit{neighbour\_calls}\}$.
\end{example}

The distribution semantics relies on a one-to-one mapping between possible worlds and models: many PLP frameworks, e.g. ProbLog, PRISM and CP-Logic, restrict the class of valid inputs to programs where each total choice corresponds to a two-valued well-founded model~\citep{DBLP:journals/jacm/GelderRS91}. The underlying assumption is that choices are modelled exclusively by means of probabilistic facts, and that they are independent.
Under this assumption many algorithms and reasoning techniques have been developed to perform inference and parameter learning tasks, which we present in the rest of this section.

\paragraph{Inference.} The MARG inference task is the task of computing the probability of success of a query, that is, the sum of the probability of each possible world where the query is true, under the (possibly empty) given evidence. Evidence is a set of atoms whose truth value is known. In particular:
\begin{definition}[The MARG Inference task]
    \textbf{Given}
    \begin{itemize}
        \item[-] A program $\mathcal{L}$: let $G$ be the set of all ground (probabilistic and derived) atoms of $\mathcal{L}$.
        \item[-] A set $E\subseteq G$ of observed atoms (evidence), along with a vector $e$ of corresponding observed truth values ($E = e$).
        \item[-] A set $Q\subseteq G$ of atoms of interest (queries).
    \end{itemize}
    \textbf{Find}
    the marginal distribution of every query atom given the evidence,
    i.e. computing $P(q\,|\,E = e)$ for each $q \in Q$.
\end{definition}
In the ProbLog2 system the probabilistic inference task is reduced to a \emph{weighted model counting problem} ($\mathit{WMC}$)~\citep{DBLP:journals/aicom/CadoliD97}.
Weighted model counting is the problem of computing the weight of a propositional logic formula $\varphi$, given a weight function $w$ that assigns a positive (real) weight value to each literal $v$ and $\lnot v$ ($v\in G$).
Let $\mathit{MOD}(\mathcal{L})$ be the set of valid interpretations for $\mathcal{L}$, the $\mathit{WMC}$ of $\varphi$ is:
\[\mathit{WMC}_{\mathcal{L}}(\varphi)= \sum_{M\in \mathit{MOD}(\mathcal{L}),M\models \varphi}\prod_{l\in M}w(l).\]
The probability $P(q|E=e)$ is computed as $\mathit{WMC}_{\mathcal{L}}(\varphi)$ where $\varphi$ is the propositional representation of $q\land E=e$. The weight function is defined as follows: for all probabilistic facts $(p:f)$, $w(f)=p$ and $w(\lnot f)=1-p$; for all atoms $a\in E$ if $a$ is true (false) $w(a)=1$, $w(\lnot a) = 0$ (resp. $w(a)=0$, $w(\lnot a) = 1$); for all remaining (logical) atoms $a$, $w(a)=w(\lnot a) =1$.

\begin{example}\label{ex:alarm_pw}
    The possible world of Example~\ref{ex:alarm} corresponding to $\{\mathit{burglary},\mathit{neighbour\_at\_home}\}$ has probability $0.1\cdot(1-0.2)\cdot 0.5= 0.04$. To find the probability of being called we sum the probabilities of all other possible worlds whose model contains $\mathit{neighbour\_calls}$, namely $\{\mathit{earthquake},\mathit{neighbour\_at\_home}\}$ and $\{\mathit{burglary},\mathit{earthquake},\mathit{neighbour\_at\_home}\}$, which are respectively $(1-0.1)\cdot 0.2\cdot 0.5= 0.09$ and $0.1\cdot 0.2\cdot 0.5= 0.01$. Therefore, $P(\mathit{neighbour\_calls})=\mathit{WMC}(\mathit{neighbour\_calls})=0.04+0.09+0.01=0.14$.
    \begin{center}
        \begin{tabular}{ l l c }
            \hline
            Possible world's choices $\omega$                                        & Model                                                     & $P(\omega)$     \\ \hline
            $\{\}$                                                                   & $\omega$                                                  & $0.36$          \\
            $\{\mathit{neighbour\_at\_home}\}$                                       & $\omega$                                                  & $0.36$          \\
            $\{\mathit{burglary}\}$                                                  & $\omega\cup\{\mathit{alarm}\}$                            & $0.04$          \\
            $\{\mathit{earthquake}\}$                                                & $\omega\cup\{\mathit{alarm}\}$                            & $0.09$          \\
            $\{\mathit{neighbour\_at\_home},\mathit{burglary}\}$                     & $\omega\cup\{\mathit{alarm}, \mathit{neighbour\_calls}\}$ & $\mathbf{0.04}$ \\
            $\{\mathit{burglary},\mathit{earthquake}\}$                              & $\omega\cup\{\mathit{alarm}\}$                            & $0.01$          \\
            $\{\mathit{earthquake},\mathit{neighbour\_at\_home}\}$                   & $\omega\cup\{\mathit{alarm}, \mathit{neighbour\_calls}\}$ & $\mathbf{0.09}$ \\
            $\{\mathit{burglary},\mathit{earthquake},\mathit{neighbour\_at\_home}\}$ & $\omega\cup\{\mathit{alarm}, \mathit{neighbour\_calls}\}$ & $\mathbf{0.01}$ \\
            \hline
        \end{tabular}
    \end{center}

\end{example}

The task of model counting is \#P-complete in general, therefore the logic program is transformed into a representation where this task becomes tractable~\citep{DBLP:conf/uai/FierensBTGR11}.
$\mathcal{L}$ is in fact transformed by means of knowledge compilation into a \emph{smooth d-DNNF} formula (deterministic Decomposable Negation Normal Form)~\citep{DBLP:conf/ecai/Darwiche04} or a \emph{sentential decision diagram}~\citep{DBLP:conf/ecsqaru/ChoiKD13}. In this paper we focus on d-DNNFs because we base our approach on a knowledge compiler to d-DNNFs for stable model counting~\citep{DBLP:conf/aaai/AzizCMS15}.
\begin{definition}
    A \emph{NNF} is a rooted directed acyclic graph in which each leaf node is labelled with a literal and each internal node is labelled with a disjunction or conjunction. A smooth \emph{d-DNNF} is an NNF with the following properties:
    \begin{itemize}
        \item Deterministic: for all disjunctive nodes the children represent formulas pairwise inconsistent.
        \item Decomposable: the subtrees rooted in two children of a conjunction node do not have atoms in common.
        \item Smooth: all children of a disjunction node use the same set of atoms.
    \end{itemize}
\end{definition}

On d-DNNFs the task of model counting becomes tractable.
The d-DNNF is further transformed into an equivalent arithmetic circuit, by replacing conjunctions and disjunctions respectively with multiplication and summation nodes, and by replacing leaves with the weight of the corresponding literals. Arithmetic circuits allow us to efficiently perform the task of $\mathit{WMC}$.

\begin{figure}[t]

    \begin{minipage}[t]{0.48\textwidth}
        \centering
        \vspace{0pt}
        \includegraphics[width=\textwidth]{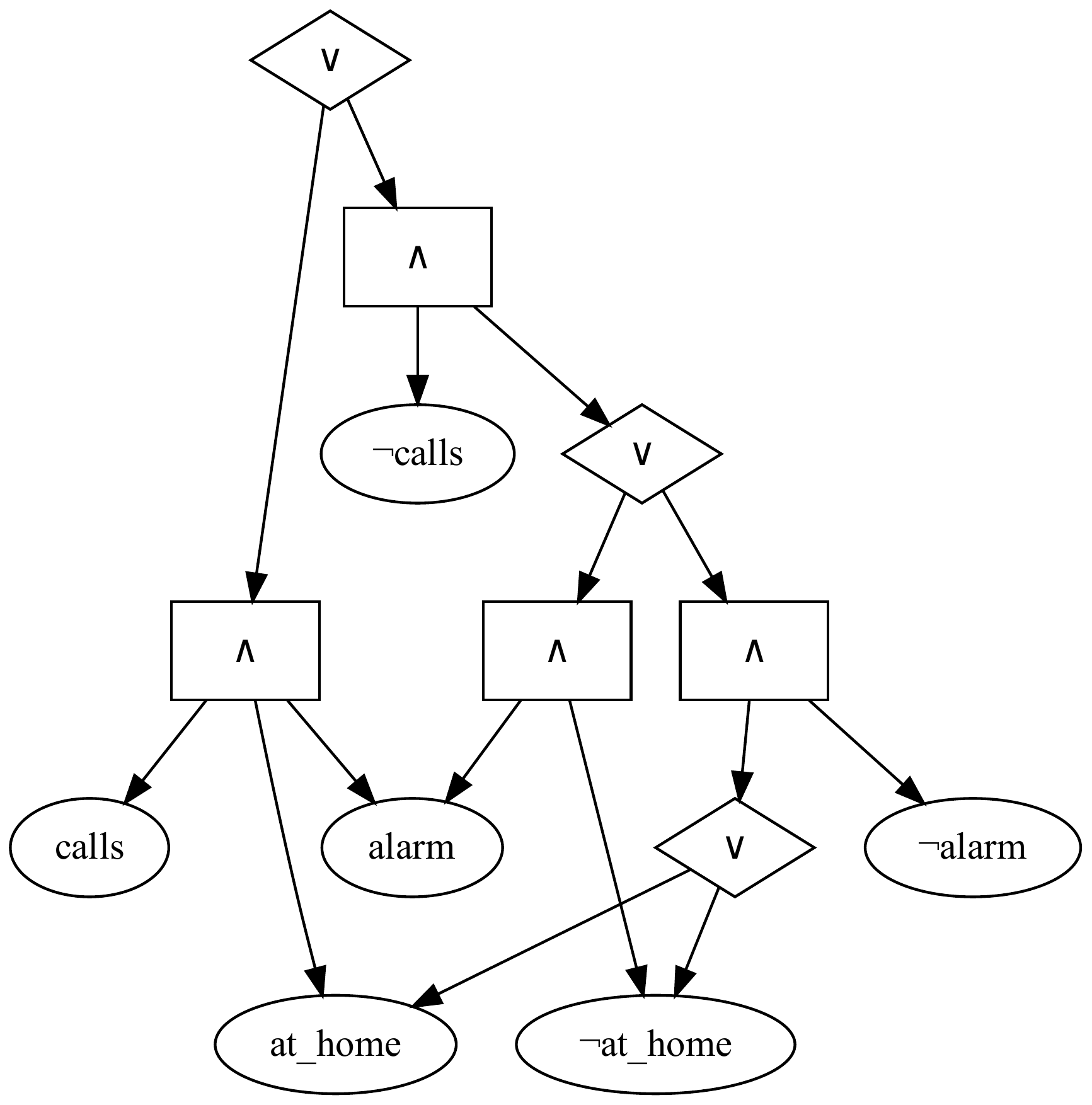}
    \end{minipage}
    \hfill
    \begin{minipage}[t]{0.48\textwidth}
        \centering
        \vspace{0pt}
        \includegraphics[width=\textwidth]{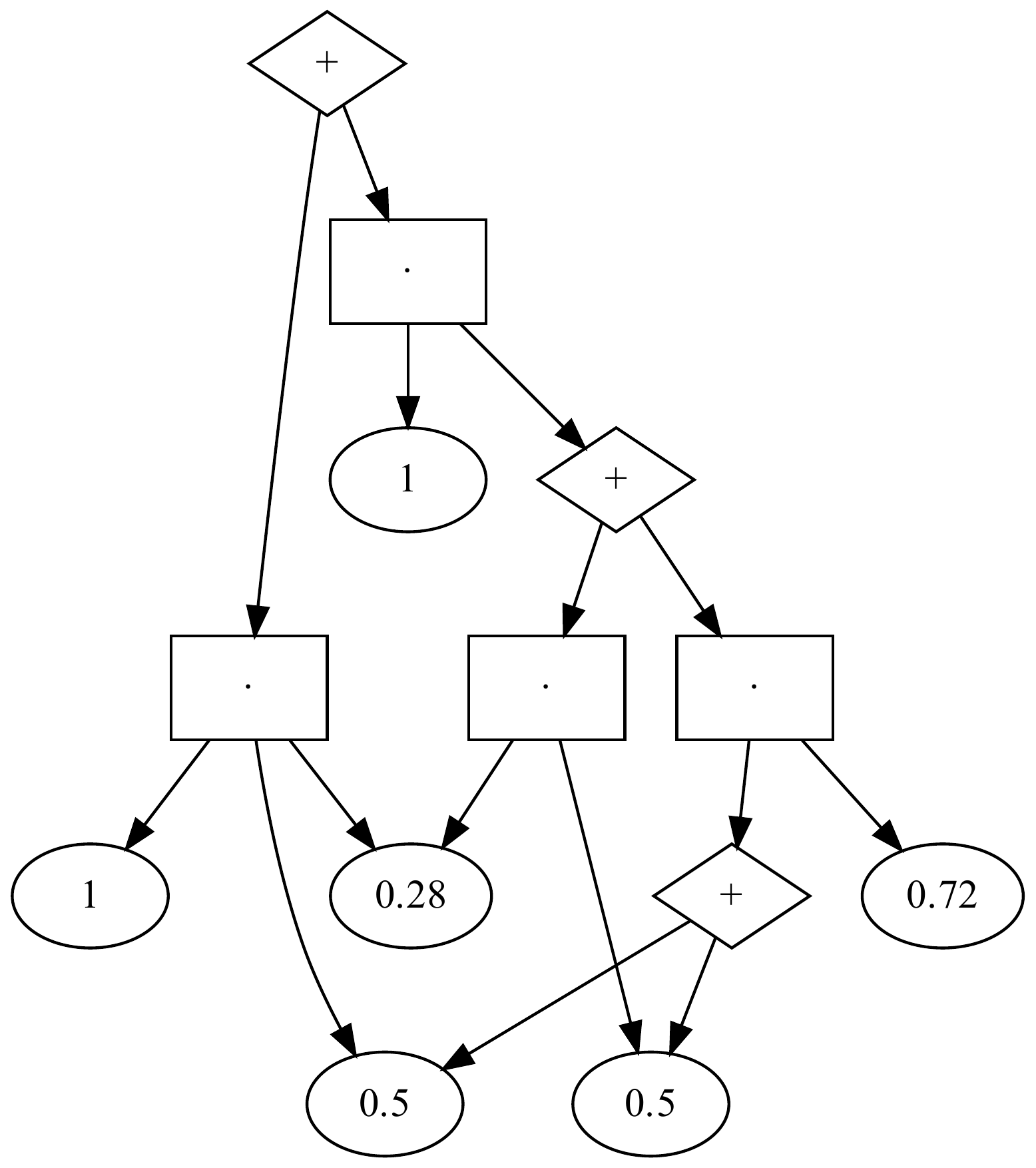}
    \end{minipage}
    \caption{Smooth d-DNNF for the implication $\mathit{neighbour\_calls} \leftarrow \mathit{alarm}, \mathit{neighbour\_at\_home}$ (left) and the corresponding arithmetic circuit (right).}
    \label{fig:kc_burglary}
\end{figure}

\begin{example}\label{ex:alarm_kc}
    Consider the rule $\mathit{neighbour\_calls} \leftarrow \mathit{alarm}, \mathit{neighbour\_at\_home}$. The corresponding smooth d-DNNF and arithmetic circuit are represented in Figure~\ref{fig:kc_burglary}. The root of the circuit represents all possible assignments to the variables. The leftmost and-node corresponding to the assignment $\mathit{calls}$ corresponds in the arithmetic circuit to the weight $0.14$, the probability of it being true. Vice versa, the top and-node, corresponding to the assignment $\mathit{\lnot calls}$, represents in the arithmetic circuit the weight $0.86$.
    We simplified Example~\ref{ex:alarm} by considering $\mathit{alarm}$ a probabilistic fact with probability $0.28$, which is the probability that can be inferred from $\mathit{burglary}$ and $\mathit{earthquake}$ (cfr. Example~\ref{ex:alarm_pw}).
\end{example}

\paragraph{Learning.} Learning from interpretations of parameters in a ProbLog program is implemented in a likelihood maximization setting~\citep{DBLP:conf/pkdd/GutmannTR11}.
In the \emph{learning from interpretations} task the input is a program $\mathcal{L}$ where some probabilities are unknown (parameters), and a set of interpretations for $\mathcal{L}$.
Interpretations can be partial, that is, the observation of truth values contains only a subset of all atoms.
The goal is to estimate the value of the parameters such that the predicted values maximize the likelihood of the given interpretations:
\begin{definition}[Max-Likelihood Parameter Estimation]
    \textbf{Given}
    \begin{itemize}
        \item[-] A program $\mathcal{L}\mathbf{(p)} = F {\cup} R$ where $F$ is a set of probabilistic facts and $R$ contains the rules describing the background knowledge. $\mathbf{p} = \langle p_1, ..., p_N\rangle$ is a set of unknown parameters attached to probabilistic facts.
        \item[-] A set $\mathbb{I}$ of (partial) interpretations $\{I_1, ..., I_M\}$ as training examples.
    \end{itemize}
    \textbf{Find}
    the maximum likelihood probabilities $\widehat{\mathbf{p}} = \langle \widehat{p_1}, ..., \widehat{p_N}\rangle$ for the interpretations in $\mathbb{I}$. Formally,
    \[
        \widehat{\mathbf{p}} = \arg \max_\mathbf{p} P(\mathbb{I}|\mathcal{L}\mathbf{(p)}) = \arg\max_\mathbf{p} \prod^M_{m=1} P(I_m|\mathcal{L}\mathbf{(p)}).
    \]
\end{definition}

Parameters are iteratively updated by an alternation of an expectation step (E step) which computes the expected value of the learnable parameters under the current model, and the maximization step (M step) which updates the parameters estimates to maximize the likelihood of the observed data.
\begin{example}\label{ex:base_learn}
    Consider Example~\ref{ex:alarm_kc} but now the probability of $\mathit{alarm}$ is a \emph{learnable parameter}. In ProbLog the expression $\mathit{t(0.4)::alarm.}$ denotes a learnable probability which is initialized at $0.4$ in the EM algorithm:
    \begin{program}
        \begin{array}{l}
            t(0.4)::alarm.                    \\
            0.5::at\_home.                    \\
            calls \leftarrow alarm, at\_home. \\
        \end{array}
    \end{program}
    Given a set of interpretations reflecting the original probability distribution we can learn the probability of $\mathit{0.28::alarm.}$ even if the observation of alarm is not given. In fact, we provide three types of partial interpretations:
    \begin{enumerate}
        \item[$I_1$:] $\{\lnot calls,\lnot at\_home\}$ (50 examples)
        \item[$I_2$:] $\{\mathit{calls}, \mathit{at\_home}\}$  (14 examples)
        \item[$I_3$:] $\{\mathit{\lnot call},  at\_home\}$ (36 examples)
    \end{enumerate}
    The first corresponds to the possible worlds where $\mathit{at\_home}$ prevents the call regardless of $\mathit{alarm}$ ($50\%$ of the examples corresponding to the probability of $\mathit{at\_home}=\mathit{false}$). The second is the case where both are true and the third where $\mathit{alarm}$ is false.
    The first iteration of the EM algorithm considers the frequency of each example and the probability according to the model with the probability of $\mathit{alarm}$ initialized at $0.4$.
    The iteration 0 then associates to $I_1$ the current probability of $\mathit{alarm}$ being either true or false, that is, $0.4$. On the other hand, in $I_2$ (resp. $I_3$) $\mathit{alarm}$ has probability $1$ (resp. $0$) of being true (E step). The updated probability that maximizes the probability of the set of training examples is thus $\frac{50\cdot 0.4 + 14\cdot 1 + 36\cdot 0}{100}=0.34$ (M step). Iteration 1 then repeats the E step with the updated probability $0.34::\mathit{alarm}$. The probabilities of $\mathit{alarm}$ in $I_1,I_2$ and $I_3$ are respectively $0.34,1.0$ and $0$ (E step) and the new updated probability is $\frac{50\cdot 0.34 + 14\cdot 1 + 36\cdot 0}{100}=0.31$ (M step). Further iterations converge to the true probability of $0.28::\mathit{alarm}$.
\end{example}

\paragraph{Causal effects.} An alternative view of ProbLog that we exploit in this paper is in terms of CP-logic, a logical language for representing probabilistic causal laws. Probabilistic causal laws model the probability distribution of a set of random variables that are related by a causal process, that is, the variables interact through a sequence of non-deterministic or probabilistic events. ProbLog and CP-logic are closely related as the syntax of a ProbLog program is similar to that of a CP-theory in which each rule has precisely one head atom. Moreover, the semantics of a ground ProbLog program coincides completely with the semantics of CP-Logic. This close relation makes available in ProbLog a modelling technique from CP-logic that we exploit in this paper: the \emph{inhibition effect}~\citep{DBLP:conf/pgm/MeertV14} and \emph{negations in the head}~\citep{DBLP:journals/corr/Vennekens13}.

Negation in the head gives an interpretation of the negation of heads from Answer Set Programming in the context of epistemic reasoning and logic theories defining causal mechanisms. In the rest of the paper we adopt this interpretation for negated heads, where the goal is to describe epistemic causal effects and not to obtain classical negation from negation as failure.~\cite{DBLP:journals/corr/Vennekens13} discusses in detail the differences with classic negation in Answer Set Programming.
In the context of PLP, each rule $\lnot h\leftarrow b_1,\dots,b_n.$ is interpreted by replacing all heads $h$ in the program with a new atom $h_{pos}$ and all heads $\lnot h$ with a new atom $h_{neg}$, and the rule $h\leftarrow h_{pos},{\sim}h_{neg}.$ is added. This last rule thus defines that $h$ can be inferred from any of the causes of $h$ (making $h_{pos}$ true) only if there is no cause for believing in $h_{neg}$, the opposite of $h$. The inhibition effect is the resulting decrease in the probability of $h$ when the probability of $\lnot h$ ($h_{neg}$) increases.

The \emph{noisy-or} effect~\citep{DBLP:conf/pgm/MeertV14} describes the aggregation of the causal probabilities for an atom which appears as the head of multiple (causal) rules.

\begin{example}
    We adapt the travel example from~\citep{DBLP:conf/pgm/MeertV14} to describe the inhibition effect in ProbLog. Alice is travelling during a pandemic and therefore there is a chance of getting infected. This probability is influenced by two factors: first, if Alice travels with public means of transport there is some probability of being infected there. Second, if Alice is vaccinated then there is a chance that she will not be infected. We encode this last rule with the inhibition effect, which describes a decrease in the probability of infection when vaccinated.
    \begin{program}
        \begin{array}{l}
            0.2::\mathit{riskyTravel}(alice).                                     \\
            0.7::\mathit{vaccinated}(alice).                                      \\
            0.4::\mathit{infected(alice)} \leftarrow \mathit{riskyTravel}(alice). \\
            0.33::\lnot \mathit{infected(alice)} \leftarrow vaccinated(alice).
        \end{array}
    \end{program}
    The two rules correspond to the following rewriting:
    \begin{program}
        \begin{array}{l}
            0.4::\mathit{infected\_pos(alice)} \leftarrow riskyTravel(alice). \\
            0.33::\mathit{infected\_neg(alice)} \leftarrow vaccinated(alice). \\
            \mathit{infected(alice)}\leftarrow \mathit{infected\_pos(alice)}, {\sim}\mathit{infected\_neg(alice)}.
        \end{array}
    \end{program}
\end{example}

\paragraph{Stable Models.} In this paper we will extend ProbLog's semantics, based on two-valued unique well-founded models, with stable model semantics~\citep{DBLP:conf/iclp/GelfondL88}, where models are two-valued, but a logic program can have more than one stable model.
Given a normal logic program $\mathcal{L}$ and a set $S$ of atoms interpreted as $true$, called \emph{candidate model}, stable model semantics is defined in terms of the \emph{reduct} of a program $\mathcal{L}$ w.r.t. $S$. The \emph{reduct} of $\mathcal{L}$ w.r.t. $S$, $\mathcal{L}^S$, is defined as follows:
1) for all $r\in \mathcal{L}$ remove $r$ from $\mathcal{L}$ if $a\in S$ is negated in the body of $r$;
2) for all $r\in \mathcal{L}$ remove from the body of $r$ all $\lnot a$ if $a\notin S$.
If $S$ is a minimal model for $\mathcal{L}^S$ then $S$ is a \emph{stable model} (\emph{answer set}) for $\mathcal{L}$.

\subsection{Probabilistic Argumentation}
\begin{figure}[t]
    \centering

    \begin{tikzpicture}[->,>=stealth',shorten >=1pt,auto,node distance=2.8cm,
            semithick, every text node part/.style={align=center}]
        \tikzstyle{every state}=[fill=white,shape=rectangle]

        \node[state]         (A)                    {$a_1$: Diane should stay \\ at
            hotel X for her holidays.};
        \node[state]         (B) [right=0.7cm of A]       {$a_2$: Diane should stay \\ at
            hotel Y for her holidays.};
        \node[state]         (C) [below=1cm of A] {$a_3$: Hotel X is a bit expensive};
        \node[state]         (D) [below=1cm of B]  {$a_4$: Hotel Y is close to \\ the
            street and might be noisy.};

        \path (A) edge [bend left]  node {} (B);
        \path (B) edge [bend left]  node {} (A);
        \path (C) edge []  node {} (A);
        \path (D) edge []  node {} (B);
    \end{tikzpicture}
    \caption{Abstract argumentation framework. Edges represent attacks, nodes are arguments. }
    \label{fig:basic_arg_graph}

\end{figure}
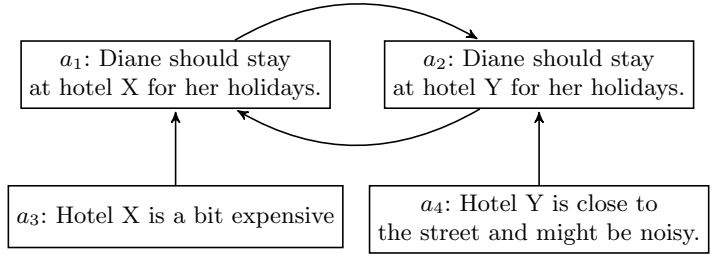

We will consider the application of PLP and \smp{} to probabilistic argumentation problems~\cite{hunter2021probabilistic}.
An \emph{abstract argumentation framework} (or \emph{argument graph})~\citep{DBLP:journals/ai/Dung95} is a pair $(A,R)$ where $A$ is a set of arguments and $R\subseteq A\times A$ is a binary (attack) relation over $A$. Figure~\ref{fig:basic_arg_graph} represents an example from~\cite{DBLP:journals/ai/HunterPT20} of argument graph $(A,R)$ where  $A=\{a_1,a_2,a_3,a_4\}$ and $R=\{(a_1,a_2),(a_2,a_1),(a_3,a_1),(a_4,a_2)\}$.

In recent years several extensions and modifications of this formulation have been proposed in order to encode more complex argumentation reasoning systems. Among the most relevant, we find \emph{bipolar} argument frameworks~\citep{DBLP:conf/nmr/AmgoudCL04}, \emph{weighted} argumentation frameworks~\citep{DBLP:conf/ijcai/AmgoudBDV17}, and \emph{probabilistic} argumentation frameworks~\citep{DBLP:conf/tafa/LiON11}.

Bipolar argumentation frameworks introduce a \emph{support} relation $R^+$ along with the attack relation $R^-$. A bipolar argument graph is thus a triple $(A,R^-,R^+)$, where $A$ is a set of arguments and $R^-\subseteq A\times A$ (resp. $R^+\subseteq A\times A$) is a binary attack (support) relation over $A$.

Weighted argumentation frameworks $(A,R,w)$ augment the traditional argument graph $(A,R)$ with a weighting function $w: A\rightarrow[0,1]$ and belong to a general class of \emph{gradual} argumentation frameworks~\citep{DBLP:journals/jair/CayrolL05}, where arguments' acceptability is evaluated on a fine-grained numerical scale or ranking.

Gradual and bipolar frameworks are combined into \emph{Quantitative Bipolar Argumentation Frameworks} (QBAFs)~\citep{DBLP:journals/ijar/BaroniRT19}, where each argument has a (possibly empty) set of attackers, a (possibly empty) set of supporters, and an initial evaluation (possibly the same for all arguments) on a chosen scale. These elements contribute to a final argument evaluation, provided by a strength function on the chosen scale. A QBAF is thus a quadruple $(A,R^-,R^+,\tau)$  consisting of a finite set $A$ of arguments, a binary (attack) relation $R^-$ on $A$, a binary (support) relation $R^+$ on $A$ and a total function $\tau: A \rightarrow I$, where $I$ is the chosen scale. $\tau(a)$ is thus the base score of $a$, for any $a \in A$.

\begin{figure}[b]
    \centering

    \begin{tikzpicture}[->,>=stealth',shorten >=1pt,auto,node distance=2.8cm,
            semithick]
        \tikzstyle{every state}=[fill=white,shape=rectangle]

        \node[state]         (A)                    {$a_1$:0.6};
        \node[state]         (B) [right of=A]       {$a_2$:0.7};
        \node[state]         (C) [left of=A] {$a_3$:0.4};
        \node[state]         (D) [right of=B]  {$a_4$:0.2};

        \path (A) edge [bend left]  node {0.7} (B);
        \path (B) edge [bend left]  node {0.4} (A);
        \path (C) edge []  node {0.9} (A);
        \path (D) edge []  node {0.8} (B);
    \end{tikzpicture}
    \caption{Probabilistic abstract argumentation framework. Edges represent attacks, nodes are arguments (cfr. Figure~\ref{fig:basic_arg_graph}). }
    \label{fig:basic_prob_arg_graph}

\end{figure}
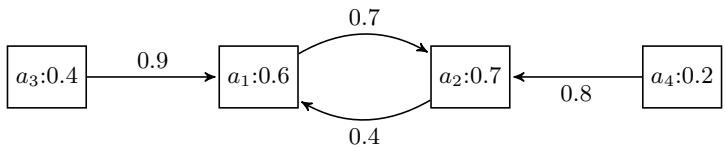

Finally, a \emph{probabilistic argument graph}~\citep{DBLP:conf/tafa/LiON11} is a tuple $(A,R,P_A,P_{R})$ where: $(A,R)$ is an argument graph and $P_A$ and $P_{R}$ are functions: $P_A:A\rightarrow[0,1]$ and $P_{R}:A\times A\rightarrow[0,1]$. Figure~\ref{fig:basic_prob_arg_graph} describes a probabilistic argument graph derived from the example in Figure~\ref{fig:basic_arg_graph}.
The values assigned by the functions $P_A$ and $P_R$ are usually interpreted with two different approaches: the \emph{constellations} approach and the \emph{epistemic} approach~\citep{DBLP:journals/ijar/Hunter13}.
The constellations approach interprets the probabilities as uncertainty over the structure of the graph and models a probability distribution over subgraphs. Each subgraph is evaluated according to the traditional extension-based semantics~\citep{DBLP:journals/ai/Dung95}. An extension is a subset of the arguments that is acceptable w.r.t. a given criterion, therefore a distribution over acceptable arguments is derived from 1) the probability distribution over the possible subgraphs, and 2) the subset of acceptable arguments in each subgraph.
The \emph{epistemic} approach on the other hand interprets the probabilities as a direct measure of an agent's belief in the arguments. The accepted arguments are those having probability higher than 0.5, and the argument graph is used to reason about the consistency of the agent's beliefs with respect to the relations encoded in the graph.

\section{Modelling Probabilistic Argumentation}\label{sec:args}

Let us now answer the question as to whether PLP can effectively model argumentation processes and  reason over their intrinsic uncertainty. Our answer takes the form of encoding argumentation problems and their uncertainty in probabilistic logic programs. By encoding arguments and uncertainty using PLP, we embed argumentation in probabilistic approaches to automated reasoning under uncertainty.  More specifically,  probabilistic logic programs represent the joint probability distribution. Modelling the joint probability distribution is central to the success of probabilistic graphical models~\citep{DBLP:journals/ker/Parsons11a}, which are the most popular tools for learning and reasoning about uncertainty in AI.   Naively specifying joint probability distribution however requires a number of parameters exponential in the number of random variables~\citep{DBLP:books/daglib/0066829, DBLP:journals/aim/Charniak91}.  The success of probabilistic graphical models and Bayesian Networks~\citep{DBLP:books/daglib/0066829} is based on the use of a graphical structure that represents a set of conditional independence assumptions, which allow to compactly encode the joint probability distributions in a factorized form. In Bayesian Networks, for instance, the key assumption is that each variable is conditionally independent of its non-descendants, given its parents.  Two random variables $X$ and $Y$ are said to be  \emph{conditionally independent given} $Z$, a third variable, if in a joint probability distribution $P$ including the three variables $P(X|Y,Z) = P(X|Z)$ whenever $P(Y,Z)>0$, that is, $Y$ does not provide extra information  $X$, if we already know $Z$.  In argumentation, conditional probabilities such as $P(X|Z)$ can be used to update the belief in $X$ in the light of the argument Z, while conditional independencies can  be used to reason about the (ir)relevance of a certain argument $Y$ in the context of $P(X|Z)$.

By approaching probabilistic argumentation problems from a PLP perspective, we apply the principles in modelling uncertainty in AI formulated by~\cite{DBLP:books/daglib/0066829}:
\begin{quotation}
    We will also stress that probability theory is unique in its ability to process context-sensitive beliefs,
    and what makes the processing computationally feasible is that the information
    needed for specifying context dependencies can be represented by graphs and
    manipulated by local propagation.
\end{quotation}
This suggests an interpretation of the argument graph as a probabilistic graphical model, that is, a graph modelling the conditional independencies between random variables (arguments). However, Bayesian Networks are directed acyclic graphs, and with argument graphs we are concerned with more complex relations involving cycles and negations (attacks). This motivates the need for a richer framework, PLP, and semantics (Section~\ref{sec:semantics}) to derive the joint probability distribution.
The argument graph thus expresses how conditional independencies and context dependencies shape a joint belief distribution across the arguments.
We thus follow the epistemic interpretation of probabilities where probabilities measure the degree of belief in the arguments.
As \cite{DBLP:journals/ijar/Hunter13} remarks, the emphasis in the epistemic approach is to find a probability distribution that is rational with respect to the argument graph.
With PLP, the derivation of the joint probability distribution from the argument graph mirrors the process of evaluating personal beliefs described by~\cite{nilsson2014understanding}:
\begin{quotation}
    If we were to examine the relationships among all of our beliefs carefully, an impossible task in practice but one
    that is interesting to think about, we would see that some of them should make others more credible and some less.
    They would even compete among themselves with conflicting influences. We can imagine all of the beliefs in our large network of beliefs ``fighting it out'' to agree finally on the strength of each belief in the network. When they do finally agree, we say that the beliefs \emph{cohere}.
\end{quotation}

We begin by defining the type of models that describe a probabilistic argumentation problem, that is, we encode the problem in a probabilistic logic program. Following the epistemic interpretation of probabilities we map arguments to logical atoms whose probability (belief) can be queried in the traditional PLP style, i.e. querying marginal  and conditional probabilities (beliefs) given evidence. Therefore, we do not define an additional layer of logic rules specifying  argumentation semantics over a (probabilistic) graph, as in the constellations approach.
Mapping a probabilistic argument graph to a probabilistic logic program means defining its semantics, because the PLP semantics of choice for the program in turn defines the graph's semantics. For this reason, we first discuss the informal semantics of the mapping, and then characterize its formal semantics by means of PLP (Section~\ref{sec:semantics}). In the following sections we describe in detail a PLP system that is capable of reasoning over such semantics.

\paragraph{Mapping.}
Given a probabilistic argument graph $G=(A,R^-,P_A,P_{R^-})$, we associate each argument $a\in A$ to a random variable $arg(a)$ representing the belief in $a$. This belief is described by means of the two probability functions: $P_A$ and $P_{R^-}$. We interpret $P_A(a)$ as the prior belief for $a$, that is, a bias independent of the other arguments, similar to a base score in a QBAF. The independence assumption of the prior beliefs in the arguments does not mean that arguments are assumed to be independent, but only the degree to which the agent is biased towards them. The \emph{posterior} belief in an argument is determined by a combination of the bias and the attacks received.
We interpret an attack $P_{R^-}(a,b)$ as the belief that accepting $a$ causes the rejection of $b$. Since we measure beliefs by means of random variables, the natural interpretation of attacks is that the belief in the target is (negatively) conditioned by the belief in the attacker, therefore believing in $a$ causes a decrease in the belief of $b$.
Therefore, the belief in an argument depends on the belief in the attacking arguments: a strong belief in any of the attackers can cause the agent to not believe the attacked argument.
In Section~\ref{sec:background} we showed how in PLP such causal relation is expressed by means of the \emph{inhibition effect}, therefore we model attacks with negation in the heads.\newpage

\begin{definition}
    Given a probabilistic argument graph $G=(A,R^-,P_A,P_{R^-})$ the corresponding probabilistic logic program $\mathcal{L}$ is a program where:
    \begin{itemize}
        \item for each $a\in A$ and $P_A(a)=p$:
              \begin{itemize}
                  \item[\labelitemi] $p::bias(a). \in \mathcal{L}$
                  \item[\labelitemi]  $arg(a)\leftarrow bias(a).\in \mathcal{L}$
              \end{itemize}
        \item for each $(a,b)\in R^-$ and $P_{R^-}((a,b))=p$:
              \begin{itemize}
                  \item[\labelitemi] $p::\lnot arg(b) \leftarrow arg(a).\in \mathcal{L}$
              \end{itemize}
    \end{itemize}
\end{definition}

\begin{example}\label{ex:arglp_base}
    Consider the example in Figure~\ref{fig:basic_prob_arg_graph}: the corresponding program is:
    \begin{program}
        \begin{array}{ll}
            0.6::bias(a_1).                          & 0.7::bias(a_2).                          \\
            0.4::bias(a_3).                          & 0.2::bias(a_4).                          \\
            arg(a_1)\leftarrow bias(a_1).            & arg(a_2)\leftarrow bias(a_2).            \\
            arg(a_3)\leftarrow bias(a_3).            & arg(a_4)\leftarrow bias(a_4).            \\
            0.9::\lnot arg(a_1) \leftarrow arg(a_3). & 0.8::\lnot arg(a_2) \leftarrow arg(a_4). \\
            0.4::\lnot arg(a_1) \leftarrow arg(a_2). & 0.7::\lnot arg(a_2) \leftarrow arg(a_1).
        \end{array}
    \end{program}
\end{example}

A rule $\lnot arg(b) \leftarrow arg(a)$ therefore says that believing argument $a$ causes not believing in $b$. In Section~\ref{sec:background} we showed how the negation in the head is interpreted in PLP by rewriting the rule as: $arg_{neg}(b) \leftarrow arg(a).$ $arg(b)\leftarrow arg_{pos}(b), {\sim} arg_{neg}(b).$ expresses ``believing $a$ causes believing in a counterargument for $b$ and $b$ cannot be believed when a counterargument is believed''.
The probability $p=P_{R^-}((a,b))$ measures the belief in the effectiveness of the attack, that is, how likely it is to conclude that argument $b$ is rejected given that $a$ is believed.

\begin{example}\label{ex:arglp_base_rw}
    The rules in Example~\ref{ex:arglp_base} are internally rewritten as follows.
    \begin{program}
        \begin{array}{ll}
            0.6::bias(a_1).                                            & 0.7::bias(a_2).                          \\
            0.4::bias(a_3).                                            & 0.2::bias(a_4).                          \\
            arg_{pos}(a_1)\leftarrow bias(a_1).                        & arg_{pos}(a_2)\leftarrow bias(a_2).      \\
            arg_{pos}(a_3)\leftarrow bias(a_3).                        & arg_{pos}(a_4)\leftarrow bias(a_4).      \\
            0.9::f_1.                                                  & 0.4::f_2                                 \\
            0.8::f_3                                                   & 0.7::f_4                                 \\
            arg(a_1) \leftarrow arg_{pos}(a_1), {\sim} arg_{neg}(a_1). & arg_{not}(a_1) \leftarrow arg(a_3), f_1. \\
            arg(a_2) \leftarrow arg_{pos}(a_2), {\sim} arg_{neg}(a_2). & arg_{not}(a_1) \leftarrow arg(a_2),f_2.  \\
            arg(a_3) \leftarrow arg_{pos}(a_3), {\sim} arg_{neg}(a_3). & arg_{not}(a_2) \leftarrow arg(a_4),f_3.  \\
            arg(a_4) \leftarrow arg_{pos}(a_4), {\sim} arg_{neg}(a_4). & arg_{not}(a_2) \leftarrow arg(a_1),f_4.  \\
        \end{array}
    \end{program}
\end{example}

We started by considering basic probabilistic argument graphs, but PLP can encode much more complex relations over random variables, hence we exploit the flexibility and expressivity of PLP to effortlessly extend this modelling technique to incorporate many relevant aspects of argumentation in a framework capable of encoding:
\begin{itemize}
    \item Quantitative evaluations of relations.
    \item Support relations.
    \item Attack (and supports) from sets of arguments.
    \item Distinctions between proponents of arguments.
\end{itemize}
We showed in Example~\ref{ex:arglp_base} how annotated rules express a fine-grained quantification of how strong the causal relation between the arguments (random variables) is.

\paragraph{Support.} Support relations are symmetric to attacks: we consider the (bipolar) extension of a probabilistic argument graph with a relation support $R^+$ ($R^+\subseteq A\times A, R^+\cap R^-=\emptyset$) and the corresponding probability $P_{R^+}:R^+\rightarrow [0,1]$. Believing in a supporter $a$ causes believing in $b$ with probability $p$: this corresponds to an increase of the belief in the supported argument $b$.
This relation is encoded as a simple inference rule:

\begin{definition}
    Given a bipolar probabilistic argument graph $G=(A,R^-,R^+,P_A,P_{R^-},P_{R^+})$ the corresponding probabilistic logic program $\mathcal{L}$ is a program where:
    \begin{itemize}
        \item for each $a\in A$ and $P_A(a)=p$:
              \begin{itemize}
                  \item[\labelitemi] $p::bias(a). \in \mathcal{L}$,
                  \item[\labelitemi]  $arg(a)\leftarrow bias(a).\in \mathcal{L}$;
              \end{itemize}
        \item for each $(a,b)\in R^-$ and $P_{R^-}((a,b))=p$:
              \begin{itemize}
                  \item[\labelitemi] $p::\lnot arg(b) \leftarrow arg(a).\in \mathcal{L}$;
              \end{itemize}
        \item for each $(a,b)\in R^+$ and $P_{R^+}((a,b))=p$:
              \begin{itemize}
                  \item[\labelitemi] $p:: arg(b) \leftarrow arg(a).\in \mathcal{L}$.
              \end{itemize}

    \end{itemize}
\end{definition}

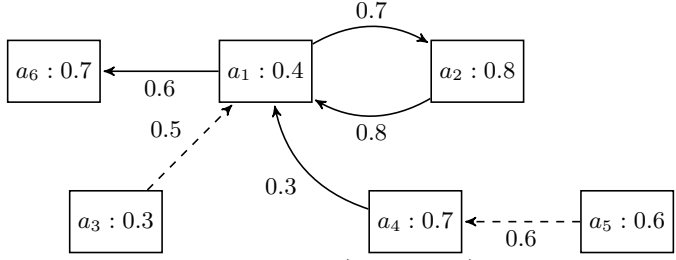
\begin{figure}
    \centering

    \begin{tikzpicture}[->,>=stealth',shorten >=1pt,auto,node distance=2.8cm,
            semithick]
        \tikzstyle{every state}=[fill=white, shape=rectangle]

        \node[state]         (A)                    {$a_1:0.4$};
        \node[state]         (B) [right of=A]       {$a_2:0.8$};
        \node[state]         (C) [below left of=A] {$a_3:0.3$};
        \node[state]         (D) [below right of=A]  {$a_4:0.7$};
        \node[state]         (E) [right of=D]       {$a_5:0.6$};
        \node[state]         (F) [left of=A]       {$a_6:0.7$};

        \path (A) edge [bend left]  node {0.7} (B);
        \path (B) edge [bend left]  node {0.8} (A);
        \path (C) edge [dashed]  node {0.5} (A);
        \path (D) edge [bend left]  node {0.3} (A);
        \path (E) edge [dashed]  node {0.6} (D);
        \path (A) edge []  node {0.6} (F);
    \end{tikzpicture}
    \caption{Example~\ref{ex:running} representation. Dashed (resp. solid) edges represent supports (resp. attacks) $R^+$ (resp. $R^-$). Nodes  (resp. edges) are labelled with the corresponding bias (belief). }
    \label{fig:arg_graph}

\end{figure}

We have thus introduced all the elements required to describe Example~\ref{ex:microtext}:
\begin{example}\label{ex:running}
    Consider Example~\ref{ex:microtext}: we model $G=(A,R^-,R^+,P_{R^-},P_{R^+})$ (see Figure~\ref{fig:arg_graph}) with arguments $a_i\in A$, attacks \sloppy$R=\{(a_1,a_6),(a_4,a_1)\}$,$(a_1,a_2),(a_2,a_1)\}$, and supports $S=\{(a_5,a_4), (a_3,a_1)\}$.
    We also define the following probability functions to derive an example of a probabilistic argumentation problem: $P_A(a_1)=0.4$, $P_A(a_2)=0.8$, $P_A(a_3)=0.3$, $P_A(a_4)=0.7$, $P_A(a_5)=0.6$, $P_A(a_6)=0.7$, $P_{R^-}(a_1,a_6)=0.6$, $P_{R^-}(a_2,a_1)=0.8$, $P_{R^-}(a_1,a_2)=0.7$, $P_{R^-}(a_4,a_1)=0.3$, $P_{R^+}(a_5,a_4)=0.6$,  $P_{R^+}(a_3,a_1)=0.5$.
\end{example}
Which we can thus encode as:
\begin{example}\label{ex:arglp}
    Example~\ref{ex:running} is encoded in the following probabilistic logic program:
    \begin{program}
        \begin{array}{ll}
            \multicolumn{2}{l}{0.4::bias(a_1). \quad 0.8::bias(a_2).}                           \\
            \multicolumn{2}{l}{0.3::bias(a_3).\quad 0.7::bias(a_4). }                           \\
            \multicolumn{2}{l}{0.6::bias(a_5). \quad 0.7::bias(a_6).}                           \\
            arg(A)\leftarrow bias(A).                &                                          \\
            0.6::\lnot arg(a_6) \leftarrow arg(a_1). & 0.6::arg(a_4) \leftarrow arg(a_5).       \\
            0.3::\lnot arg(a_1) \leftarrow arg(a_4). & 0.5::arg(a_1) \leftarrow arg(a_3).       \\
            0.8::\lnot arg(a_1) \leftarrow arg(a_2). & 0.7::\lnot arg(a_2) \leftarrow arg(a_1).
        \end{array}
    \end{program}
\end{example}

\paragraph{Sets.} In this context, it is immediate to generalize relations between pairs of arguments to  relations involving multiple arguments. This is the case of argument systems~\citep{DBLP:conf/argmas/NielsenP06}, where the attack between two arguments is generalized to an attack relation from a set of arguments towards a single argument (set-attacks). We can model this in PLP by conjoining attackers (resp. supporters) in the body of rules.
\begin{example}
    $\lnot arg(a_1) \leftarrow arg(a_4), arg(a_5).$  is a causal relation effective only when all the arguments in the set of attackers (body) are believed.
\end{example}
Also in this case we can easily combine different argumentative features with each other, for example:
\begin{example}
    Examples of gradual joint relations in the setting of Example~\ref{ex:microtext}:
    \begin{itemize}
        \item gradual set attacks: $0.6::\lnot arg(a_1)\leftarrow arg(a_4),arg(a_5).$
        \item gradual set supports: $0.3::arg(a_3) \leftarrow arg(a_4), arg(a_5).$
    \end{itemize}
\end{example}

\paragraph{PLP tools.} Note that in Example~\ref{ex:arglp} we use first-order predicates to compactly define the relation between arguments and their bias. This is a modelling advantage offered by PLP, whose flexibility can also be exploited to model less standard scenarios. For instance, in the following example we distinguish between different proponents of the arguments and define the bias in terms of a measure of the trust in them:
\begin{example}
    Consider Example~\ref{ex:arglp} and the distinction between arguments coming from a proponent (\sloppy$\{a_2,a_4,a_5,a_6\}$) and those introduced by an opponent ($\{a_1,a_3\}$). We can model bias based on the source of the argument as follows:
    \begin{program}
        \begin{array}{c c}
            0.4::\mathit{proponent}.               & 0.7::\mathit{opponent}.                \\
            arg(a_1)\leftarrow \mathit{opponent}.  & arg(a_2)\leftarrow \mathit{proponent}. \\
            arg(a_3)\leftarrow \mathit{opponent}.  & arg(a_4)\leftarrow \mathit{proponent}. \\
            arg(a_5)\leftarrow \mathit{proponent}. & arg(a_6)\leftarrow \mathit{proponent}.
        \end{array}
    \end{program}
    First order logic allows for example expressing compactly that the more agents propose the same argument, the higher its bias is.
    Assume we express proponents and trust as $p_1::prop(1).\cdots p_n::prop(n).$ and $proposes(i,a)$ denotes that argument $a$ is backed by proponent $i$, then a rule $bias(A) \leftarrow proposes(P,A), prop(P).$ defines a contribution to the bias of each argument from all its proponents, denoted by each belief $p_i$.
\end{example}

We thus showed how PLP offers an expressive language that is capable to encode complex relations between arguments and their beliefs.
Ideally, we would like to be able to perform the inference and learning tasks over such programs, and apply the traditional PLP algorithms and tools to models of argumentation problems. This means answering marginal probability queries of the kind: ``what is the inferred belief in an argument given its bias and the relationships with the other arguments?'' or answering conditional queries by reasoning over evidence, i.e. ``how does the belief in an argument change if the validity or falsehood of a set of arguments is determined?''.
The natural question arising at this point is whether the syntax is paired with semantics and reasoning systems that support this type of models.
Unfortunately, the answer is negative: the assumptions that PLP semantics impose on the input program are too restrictive for a broad class of argumentation problems.
In particular, the presence of cycles through negation in a probabilistic program $\mathcal{L} = F\cup R$ usually results in a violation of the requirement of the existence of a single two-valued well-founded model for all the possible (deterministic) logic programs $\omega\cup R$, $\omega\subseteq F$. This situation is often present in the case of reciprocal (or, in general, cyclic) attacks in argument graphs.
Example~\ref{ex:enccycle} shows how arguments attacking each other results in a cyclic relation where negation is involved:

\begin{example}\label{ex:enccycle}
    In Example~\ref{ex:arglp} rules
    $0.8::\lnot arg(a_1) \leftarrow arg(a_2).$ and $0.7::\lnot arg(a_2) \leftarrow arg(a_1).$ correspond to the following rewriting of the negated heads:
    \[
        \begin{array}{l l l}
            0.7::f. & arg_{neg}(a_2) \leftarrow f, arg(a_1). & arg(a_1) \leftarrow arg_{pos}(a_1), {\sim}arg_{neg}(a_1). \\
            0.8::g. & arg_{neg}(a_1) \leftarrow g, arg(a_2). & arg(a_2) \leftarrow arg_{pos}(a_2), {\sim}arg_{neg}(a_2).
            \\
        \end{array}
    \]
    The negation in the head thus leads to a cyclic dependency, we use $\xrightarrow{\sim}$ to denote a dependency through negation, $\rightarrow$ otherwise:
    \[ arg(a_1)\xrightarrow{\sim} arg_{neg}(a_1)\rightarrow arg(a_2)\xrightarrow{\sim} arg_{neg}(a_2) \rightarrow arg(a_1)\]
    which causes the program to not satisfy the requirement of a unique two-valued well-founded model in the possible worlds where $f$, $g$, and the probabilistic facts corresponding to the supports of the two arguments are included.
\end{example}

In the Section~\ref{sec:semantics} we discuss why such programs are not captured by traditional PLP semantics, and we propose a new semantics that, on the contrary, allows us to reason over this particular class of programs.

\section{Probabilistic Semantics}\label{sec:semantics}

In this section we present a novel PLP semantics to reason over the programs obtained from the mapping of probabilistic argumentation frameworks described in Section~\ref{sec:args}. In fact, traditional PLP frameworks cannot reason over these programs because they may define cyclic dependencies through negation (Example~\ref{ex:enccycle}).
In PLP, in particular ProbLog~\citep{DBLP:conf/uai/FierensBTGR11}, the requirement of a single two-valued well-founded model is always satisfied when the program does not define cyclic relations through negation (a sufficient but not necessary condition). In fact, in the case of well-founded semantics for normal logic programs, cycles through negation can result in a unique model where some atoms are labelled with a third value ``undefined''~\citep{DBLP:journals/jacm/GelderRS91}.
At the same time, in stable model semantics, which is two-valued, cycles through negation can lead to one, many, or zero stable models~\citep{DBLP:conf/iclp/GelfondL88}.

\begin{example}
    \citeN{DBLP:journals/jacm/GelderRS91} shows that the program $a\leftarrow {\sim}b. b\leftarrow {\sim}a.$  has no two-valued well-founded model, hence the three-valued well-founded model considers both $a$ and $b$ as \emph{undefined}. On the other hand, the program has two stable models, namely $\{a,\lnot b\}$ and $\{\lnot a, b\}$.
\end{example}

For this reason in Answer Set Programming non-deterministic choices between atoms can be expressed by recursive dependencies through negation.
The answer sets of a program are possible sets of beliefs that a rational agent may hold on the basis of the information expressed by the logic rules~\citep{gel91b}.
For this reason, we will consider stable model semantics in order to handle non-probabilistic choices introduced by the logic component.

The case with multiple stable models corresponds to the expression of non-deterministic aspects of the problem modelled by the logic rules~\citep{DBLP:conf/pods/SaccaZ90}.

\begin{example}\label{ex:hotels_sm}
    Consider Example~\ref{fig:basic_arg_graph}: if we think that Diane will find Hotel X too expensive with probability 0.6 and that Hotel Y will be noisy with probability 0.7, we can write the following program, with the equivalent rewriting on the right:
    \begin{program}
        \begin{array}{l | l}
            \textsc{program:}        & \textsc{internal rewriting:}                                               \\
            0.6::\te.                & 0.6::\te.                                                                  \\
            0.7::\tn.                & 0.7::\tn.                                                                  \\
            \hy \leftarrow \te.      & \mathit{stay\_at\_Y\_pos} \leftarrow \te.                                  \\
            \hx \leftarrow \tn.      & \mathit{stay\_at\_X\_pos} \leftarrow \tn.                                  \\
            \lnot\hy \leftarrow \hx. & \mathit{stay\_at\_Y\_neg} \leftarrow \hx.                                  \\
            \lnot\hx \leftarrow \hy. & \mathit{stay\_at\_X\_neg} \leftarrow \hy.                                  \\
                                     & \hx \leftarrow \mathit{stay\_at\_X\_pos}, {\sim} \mathit{stay\_at\_X\_neg} \\
                                     & \hy \leftarrow \mathit{stay\_at\_Y\_pos}, {\sim} \mathit{stay\_at\_Y\_neg} \\
        \end{array}
    \end{program}
    Intuitively, if the counterarguments for both hotels are considered, i.e. the possible world where $\te$ and $\tn$ are included, then we are left with a choice between hotel X and hotel Y. In fact, while the first two rules of the program suggest that Diane should stay at both hotels, the last two rules state that this is logically not possible and only one of the two should be chosen. Therefore, in this possible world logic defines a non-determisitic choice that was not modelled by probabilistic facts.
\end{example}

This case is not compatible with PLP frameworks because the underlying assumption is that uncertainty is fully captured by each total choice and therefore that a user exactly knows the causes and effects of all relevant non-deterministic events that might happen, e.g. ``believing in argument $a$ determines a choice between arguments $b$ and $c$''. However, we showed in Section~\ref{sec:args} (for instance Example~\ref{ex:arglp_base} and~\ref{ex:arglp}) that this is not always the case and obtaining such complete information over the choices may require inference itself. For instance, in argumentation some choices are induced by the relations between arguments (Example~\ref{ex:enccycle}) and may be required only if some particular conditions apply, e.g. not accepting an argument, which may depend in turn on other probabilistic aspects of the problem.
Stable model semantics provides a solution to handling non-probabilistic choices introduced by the logic component of a program. At the same time, the absence of a stable model signals that a given selection of probabilistic facts $F$ (total choice) is inconsistent with the knowledge encoded in the rules $R$ of the program.

A program with zero stable models reflects the case where it is not possible to define an interpretation of the atoms that satisfies all logic rules, therefore a program that has at least one stable model is called ``consistent''~\citep{DBLP:reference/fai/Gelfond08}.

\begin{example}\label{ex:barber}
    Consider a program modelling the famous barber paradox: in a village the barber shaves all men, and those only, who do not shave themselves, does the barber shave himself?
    \begin{program}
        \begin{array}{ll}
            0.5::\mathit{barber(bob)}. & 0.5::\mathit{villager(bob)}. \\
            \multicolumn{2}{c}{\mathit{shaves(X,Y)} \leftarrow \mathit{barber(X)}, \mathit{villager(Y)}, {\sim}\mathit{shaves(Y,Y)}.}
        \end{array}
    \end{program}
    In the possible world where \emph{bob} is both a villager and the barber (with probability $0.25$), we have an inconsistency: there is neither a two-valued well-founded model nor a stable model.
    Therefore, we will say that the program is inconsistent with probability $0.25$ and the atoms for the inconsistent possible worlds are interpreted with a third value ``inconsistent''.
    This requires us to point out a relevant consideration about the semantics of total choices.
\end{example}

In the past total choices have been defined in two different ways, as an \emph{interpretation} or a \emph{selection} of facts. They are equivalent under the assumption of a one-to-one correspondence between a total choice and the least model or the two-valued well-founded model, but this is not the case when considering a more general semantics for models.
In the original definition by~\citeN{DBLP:conf/iclp/Sato95} a total choice is an \emph{interpretation} (an assignment of truth values) of the set of probabilistic facts, whose probability distribution is extended to a distribution over least models.~\citeN{DBLP:conf/uai/FierensBTGR11}, on the other hand, define in ProbLog an atomic choice $(p:f)$ as the choice of \emph{selecting} $f$ for being included with probability $p$ or discarded with probability $1-p$. A total choice for a ProbLog program $\mathcal{L}=F\cup R$ (probabilistic facts $F$ plus rules $R$) is thus any subset $\omega\subseteq F$ of the facts, which defines a logic program $\omega\cup R$.
These two definitions are equivalent under two-valued well-founded semantics because facts are always interpreted as true. This holds also when considering programs with at least one stable model. However, if there is no stable model, it is not possible to consider a total choice an interpretation of the (probabilistic) facts.
\begin{example}
    In Example~\ref{ex:barber} defining the total choice as an interpretation assigning true to both probabilistic facts does not generalize to a normal logic program without (two-valued) models, because such interpretation is no longer a subset of any (two-valued) model. The total choice $\{\mathit{barber(bob)}, \mathit{villager(bob)}\}$ cannot be regarded as a partial interpretation where both are true because there is no model that extends such interpretation. On the contrary, considering the total choice as a selection of facts does not make any assumption on the eventual interpretation. $\{\mathit{barber(bob)}, \mathit{villager(bob)}\}$ can be thus regarded as an inconsistent choice for the given program and the logical atoms can be interpreted as inconsistent instead of \emph{true}.
\end{example}

Therefore, in this paper we adopt the definition of total choices as subsets of (probabilistic) facts, which thus makes no assumption about their interpretation in the corresponding model(s).
More specifically, our semantics uses the following principles to deal with the novel setting.

\paragraph{Zero models.} Inconsistent programs correspond to a loss of probability mass in the interpretations of probabilistic facts, because such probability mass can no longer be associated to a two-valued model.
For instance:

\begin{example}\label{ex:zero}
    An alarm goes off when a burglary or earthquake happen. The owner however thinks that the alarm was not properly installed and should not go off because of a valid reason. If the alarm does not activate then the owner is right. We can model this situation with the following (probabilistic) normal logic program:

    \begin{program}
        \begin{array}{l l}
            0.5::\bg.                      & 0.5::\eq.                            \\
            \al\leftarrow \bg.             & \al\leftarrow \eq.                   \\
            \df \leftarrow \al, {\sim}\df. & \mathit{right}\leftarrow {\sim} \al. \\
        \end{array}
    \end{program}
    There are four total choices: $\omega_1=\{\bg,\eq\}$, $\omega_2=\{\bg\}$, $\omega_3=\{\eq\}$, $\omega_4=\{\}$. The logic part of the program is inconsistent when $\al$ is true, therefore when either $\bg$ or $\eq$ are true.
    This means that for three total choices, $\omega_1,\omega_2$, and $\omega_3$, the corresponding possible world has neither a two-valued well-founded model nor stable models. In these cases the program is inconsistent with probability $P(\omega_1)+P(\omega_2)+P(\omega_3)=0.75$ and the only total choice corresponding to a consistent possible world is $\omega_4$ which corresponds to the stable model $\{\mathit{right}\}$.
\end{example}
We thus assign the probability mass associated to inconsistencies to an ``inconsistent'' state, whose probability thus quantifies how likely it is to consider an inconsistent possible world among those defined by the program. We define this in terms of a model where all atoms are labelled with a third value ``inconsistent''.
This solution has the advantage of preserving the probability mass of total choices by assigning it to the model representing an inconsistent possible world. The semantics is thus informative as to the degree of inconsistency of the logical part of the program. Note that this is different from three-valued well-founded semantics, where the atoms involved in the cycles responsible for multiple stable models are labelled with the third value (``undefined'') as well, and not all atoms are interpreted undefined at the same time.

\begin{example}
    When arguing about getting vaccinated or not there are different media we can choose to believe in: virologists, newspapers and Facebook. Virologists endorse vaccines and their safety, newspapers argue that there is some degree of risk involved, Facebook posts reject the acceptability of such degree of risk. We can model this reasoning pattern and the level of trust in the different media with the following program:
    \begin{program}
        \begin{array}{lll}
            0.9::trust\_virologists. & 0.7::trust\_newspapers. & 0.3::trust\_facebook.                                    \\
            \multicolumn{3}{l}{\mathit{safe} \leftarrow \mathit{trust\_virologists}, {\sim} \mathit{dangerous}.}          \\
            \multicolumn{3}{l}{\mathit{reasonable\_risk} \leftarrow \mathit{trust\_newspapers}, {\sim} \mathit{safe}.}    \\
            \multicolumn{3}{l}{\mathit{dangerous} \leftarrow \mathit{trust\_facebook}, {\sim} \mathit{reasonable\_risk}.} \\
        \end{array}
    \end{program}
\end{example}
The three rules determine a cyclic dependency (of odd length) with negation involved between $\mathit{safe}$, $\mathit{reasonable\_risk}$ and $\mathit{dangerous}$ and therefore trusting all media is an inconsistency, because there is no possible truth assignment to $\mathit{safe}$, $\mathit{reasonable\_risk}$, and $\mathit{dangerous}$. We will then say that this program is inconsistent with probability $0.9\cdot0.7\cdot0.3=0.189$. The remaining probability mass of $(1-0.189)=0.811$ is divided over the other possible words where the total choice of probabilistic facts is a consistent choice according to the logic of the program. These choices thus induce a two-valued assignment to the atoms of the program.

\paragraph{At least one model.} In order to give a probabilistic interpretation of logical choices, we follow the principle of maximum entropy.  The maximum-entropy principle is a widely adopted principle in Bayesian reasoning and Statistical Relational Learning. It states that the probability distribution which best represents the current state of knowledge about a system is the one with the largest entropy~\citep{Jaynes1988}. In this setting this principle translates to considering a uniform distribution for the (multiple) stable models of a total choice.
\begin{example}\label{ex:multiple}
    Consider the following (probabilistic) normal logic program:
    \begin{program}
        \begin{array}{ l l}
            0.5::\bg.                & 0.5::\eq.                 \\
            \al\leftarrow \bg.       & \df\leftarrow \eq.        \\
            \al\leftarrow {\sim}\df. & \df\leftarrow {\sim} \al.
        \end{array}
    \end{program}
    There are four total choices: $\omega_1=\{\bg,\eq\}$, $\omega_2=\{\bg\}$, $\omega_3=\{\eq\}$, $\omega_4=\{\}$. While the first three correspond to one stable model, i.e. $\mathit{MOD}(\omega_1)=\{\{\bg,\eq,\al,\df\}\}$, $\mathit{MOD}(\omega_2)=\{\{\bg,\al\}\}$, $\mathit{MOD}(\omega_3)=\{\{\eq,\df\}\}$, $\omega_4$ has two stable models: $\mathit{MOD}(\omega_4)=\{\{\al\},\{\df\}\}$.  Since this choice is not related to beliefs but rather to logical consistency, we assume that all stable models are equally probable for the given total choice $\omega_4$. In this case we will choose $\al$ half of the times and $\df$ the other half. Since $P(\omega_4)=0.25$ the probability of the model $\{\al\}$ is equal to the probability of the model $\{\df\}$, namely $0.125$.
\end{example}

\begin{example}\label{ex:hotel_sm_math}

    The program of Example~\ref{ex:hotels_sm} has four consistent possible worlds, one of which with two stable models:
    \begin{center}
        \begin{tabular}{ l c c c }
            \hline
            Possible world's choices                 & $P(\omega)$             & Model $M_\omega$                               & $\hat{P}(M_\omega)$ \\ \hline
            $\omega_1=\{\}$                          & $0.12$                  & $M_{\omega_1}=\omega_1$                        & $0.12$              \\
            $\omega_2=\{\te\}$                       & $0.18$                  & $M_{\omega_2}=\omega_2\cup\{\hy\}$             & $0.18$              \\
            $\omega_3=\{\tn\}$                       & $0.28$                  & $M_{\omega_3}=\omega_3\cup\{\hx\}$             & $\mathbf{0.28}$     \\
            \multirow{2}{*}{$\omega_4=\{\te, \tn\}$} & \multirow{2}{*}{$0.42$} & $\bigstrut M_{\omega_4}^1=\omega_4\cup\{\hy\}$ & $0.21$              \\\cline{3-4}
                                                     &                         & \bigstrut $M_{\omega_4}^2=\omega_4\cup\{\hx\}$ & $\mathbf{0.21}$     \\
            \hline
        \end{tabular}
    \end{center}
    Consider $\omega_4$: when both probabilistic facts $\te$ and $\tn$ are conjoined with the rules, there is no probabilistic justification for choosing one hotel over the other, but rules impose a choice.
    In this case, we consider the probability of $\hx$ and $\hy$ to be equivalent, because in a possible world where both counterarguments are true, a choice of a hotel is still necessary. The probabilistic part of the program which is quantified by the user is fixed in the total choice $\omega_4$, therefore in determining the models of $\omega_4$ there is no probabilistic justification for assigning one model a probability higher than another. In the definition of the probability distribution of the models $\hat{P}$, we thus divide equally the probability of the possible world determined by $\omega_4$ between its two models: $\hat{P}(M_{\omega_4}^1)=\hat{P}(M_{\omega_4}^2)=\frac{0.42}{2}=0.21$.
    The probability of an atom $a$ remains the sum of the probability of the models where $a$ is true, for example the probability of $\hx$ is $\hat{P}(M_{\omega_3}) + \hat{P}(M_{\omega_4}^2) = 0.28+0.21=0.49$.
\end{example}


\subsection{Formal semantics}
We define the models for each total choice and the corresponding probability in order to extend a probability distribution over total choices to a probability distribution over models.
Each total choice $\omega$ corresponds to: (1) a probability $P(\omega)$ as defined by distribution semantics and (2) a set
$SM(\omega)$ of stable models, over which the probability gets distributed if it is non-empty. For this distribution to be well-defined, we only consider programs with finite Herbrand base, and thus a finite number of interpretations.\newpage

\begin{definition}
    A valid \smp{} program is a probabilistic normal logic program without function symbols.
\end{definition}

\begin{definition}
    Given a valid \smp{} program $\mathcal{L}$, the probability $ P(\omega)$ of a total choice $\omega$ is:
    \[P(\omega)=\prod_{(f:p)\in\mathcal{L}, f\in\omega} p\cdot\prod_{(f:p)\in\mathcal{L}, f\not\in\omega} 1-p.\]
\end{definition}
that is, the product of the probabilities of the facts for being included/excluded from $\omega$.
\begin{example}
    The probability of the total choices in Examples~\ref{ex:zero} and~\ref{ex:multiple} is $P(\omega_1)=P(\omega_2)=P(\omega_3)=P(\omega_4)=0.5\cdot 0.5=0.25$
\end{example}
\begin{definition}
    Given a probabilistic normal logic program $\mathcal{L}=F\cup R$ and a corresponding total choice $\omega\subseteq F$, $SM(\omega)$ is the set of stable models of the (deterministic) program $\omega\cup R$.
\end{definition}

We distinguish two cases: $SM(\omega)=\emptyset$ and $|SM(\omega)|>0$, and we define accordingly a set of models for a possible world.

A model is a three-valued interpretation $(T,F)$ where $T$ is the set of true atoms and $F$ is the set of false atoms. The remaining atoms are interpreted as being part of an inconsistent possible world.
\begin{definition}
    Given a probabilistic logic program $\mathcal{L}$ and let $HB(\mathcal{L})$ be its Herbrand base. For each total choice $\omega\in \Omega_\mathcal{L}$, the corresponding set of models $\mathit{MOD}(\mathcal{L},\omega)$ is:
    \begin{equation*}
        \mathit{MOD}(\mathcal{L},\omega) =
        \begin{cases}
            \{(\emptyset, \emptyset)\}                                                           & \text{if $|SM(\omega)| = 0$.} \\
            \{(M_{\omega}, HB(\mathcal{L})\backslash M_{\omega})\;|\; M_{\omega}\in SM(\omega)\} & \text{if $|SM(\omega)| > 0$.} \\
        \end{cases}
    \end{equation*}
\end{definition}

Each total choice thus has at least one model: either the ``inconsistent'' model or one or more stable models. We extend the probability of the total choices to a distribution over the corresponding models by applying the maximum-entropy principle.

The probability of a model $M_{\omega}\in \mathit{MOD}(\mathcal{L},\omega)$ is the probability of the corresponding total choice $\omega$ normalized w.r.t. the number of the models for that possible world:
\begin{definition}
    Given a probabilistic normal logic program $\mathcal{L}$ and a total choice $\omega$, $\forall M_{\omega}\in \mathit{MOD}(\mathcal{L},\omega)$:
    \[\hat{P}(M_{\omega}) = \frac{P(\omega)}{|\mathit{MOD}(\mathcal{L},\omega)|}\]
\end{definition}

We thus derive a distribution over models $\hat{P}$ from the probability distribution over total choices $P$. We can now define the probability of the program of being inconsistent and of a query of an atom.
The degree of inconsistency of the program $\mathcal{L}$, $\mathbb{P}(\mathcal{L}\vdash\bot)$ is the sum of the probabilities of interpreting the atoms as inconsistent in each possible world:
\[\mathbb{P}(\mathcal{L}\vdash\bot)=\sum_{\omega\in\Omega_\mathcal{L}, \mathit{MOD}(\mathcal{L},\omega)=\{(\emptyset, \emptyset)\}} \hat{P}((\emptyset, \emptyset)).\]
As for the consistent possible worlds, we can query the program for the probability of an atom being interpreted as true (or false):
\begin{definition}
    Given a probabilistic normal logic program $\mathcal{L}$, the probability of success of querying $a$ is the probability of interpreting $a$ true in a consistent possible world:
    \[\mathbb{P}(a)=\sum_{a\in M_{\omega},(M_{\omega}, HB(\mathcal{L})\backslash M_{\omega})\in \mathit{MOD}(\mathcal{L},\omega),\omega\in\Omega_\mathcal{L}}\hat{P}(M_{\omega}). \]
\end{definition}
Note that we can derive the probability of an atom being false from the probability of the inconsistent possible worlds and the probability of success:
\[\mathbb{P}(\lnot a) = 1 - \mathbb{P}(\mathcal{L}\vdash\bot) - \mathbb{P}(a) \]
because the probability mass of the non-inconsistent models is divided over the two-valued stable models, therefore if an atom is in a consistent possible world, then it is interpreted either true or false.

\begin{example}
    We complete Example~\ref{ex:zero}: the probability of the program being inconsistent is $P(\omega_1)+P(\omega_2)+P(\omega_3)=0.75$. The only consistent possible world is $\omega_4$, whose only model is $\{e\}$: $\hat{P}(\rt)=0.25$, therefore $\mathbb{P}(\rt)=0.25$ and $\mathbb{P}(\lnot \rt)=0$, vice versa, $\mathbb{P}(\bg)=\mathbb{P}(\eq)=\mathbb{P}(\al)=\mathbb{P}(\df)=0$ and $\mathbb{P}(\lnot a)=\mathbb{P}(\lnot b)=\mathbb{P}(\lnot c)=\mathbb{P}(\lnot d)=0.25$.
\end{example}
\begin{example}
    We complete Example~\ref{ex:multiple}: models $\{\bg,\eq,\al,\df\}$, $\{\bg,\al\}$, $\{\eq,\df\}$ are the unique model for, respectively, $\omega_1,\omega_2,\omega_3$ hence $\hat{P}(\{\bg,\eq,\al,\df\})=\hat{P}(\{\bg,\al\})=\hat{P}(\{\eq,\df\})=P(\omega_1)=P(\omega_2)=P(\omega_3)=0.25$. The likelihood of $\omega_4$ is uniformly distributed over the two stable models $\{\al\}$ and $\{\df\}$, whose probability is thus $\frac{P(\omega_4)}{|\mathit{MOD}(\mathcal{L},\omega_4)|}=\frac{0.25}{2}=0.125$. Therefore,
    $\mathbb{P}(\bg)=\mathbb{P}(\eq)=0.5$, $\mathbb{P}(\al)=\mathbb{P}(\df)=0.625$.
\end{example}

\section{The joint distribution of beliefs}\label{sec:joint}


By defining the semantics of \smp{} programs we also define the semantics of the mapping from probabilistic argumentation problems to PLP (Section~\ref{sec:args}). Under these semantics a probabilistic argument graph thus represents a \emph{joint probability distribution} of the arguments' beliefs by means of the corresponding probabilistic logic program.
As we argued in Section~\ref{sec:args}, the joint probability distribution represents the marginal and conditional strengths of the beliefs in the arguments from their biases and relations. That is, we are concerned with determining the probabilities of logically related propositions.
\begin{example}\label{ex:marg_arg}
    In Example~\ref{ex:running} the marginal probabilities of each argument in the inferred joint probability distribution are:  $\mathbb{P}(arg(a_1))=0.29$, $\mathbb{P}(arg(a_2))=0.63$, $\mathbb{P}(arg(a_3))=0.3$, $\mathbb{P}(arg(a_4))=0.81$, $\mathbb{P}(arg(a_5))=0.6$, $\mathbb{P}(arg(a_6))=0.58$.
\end{example}

These marginal distributions are thus coherent with the initial biases and relations in the sense of~\cite{nilsson2014understanding}. \cite{DBLP:conf/sum/PolbergHT17} define several properties for epistemic argument graphs, among which coherency, to evaluate arbitrary assignments of values in $[0,1]$ under the principle that a probability higher than $0.5$ defines an argument as accepted. In the epistemic approach literature, acceptance is thus a function of the arbitrary probability values with respect to an arbitrary threshold.
Marginal probabilities, representing the agent's beliefs, are thus directly assigned to arguments, and they are evaluated with respect to local properties. \cite{DBLP:journals/corr/abs-1802-07489} generalize this approach with \emph{epistemic graphs}, where, instead of considering a single assignment, a set of belief assignments is associated to the argument graph such that they satisfy a set of constraints.  This does not narrow down the belief distribution to a single joint probability distribution, but admits a set of possible assignments compatible with the constraints.

Therefore, past work evaluates arbitrary belief assignments with respect to an arbitrary threshold: considering a belief \emph{absolute} value of 0.5 the discriminant between accepted and rejected arguments. This poses a well-known question in the PLP literature, ``Where do the numbers come from?''~\citep{DBLP:journals/aim/Charniak91}, in regard of evaluating subjective beliefs with a fixed value of reference.
The main difference with our approach is that in our framework acceptance is a function of logic and its (joint) probability is the associated (joint) belief measure. The joint belief (probability) distribution represents how the prior beliefs (biases) interact with each other (arguments' logical relations) to define the \emph{unique} joint belief distribution that is coherent with the given argumentative structure (argument graph).
The joint belief distribution defines how the \emph{relative} strength of the arguments changes when their logical relations are considered, regardless of an acceptance threshold. This expresses two principles of modelling uncertainty in AI. First, as~\cite{roma1931sul} remarks in his discussion on subjective beliefs, rather than evaluating the absolute degree of beliefs, an agent is coherent when the relative differences between such degrees do not contradict probability axioms. Second, we apply the approach of probabilistic graphical models to probability and reasoning: ``Probability is not really about numbers; it is about the structure of reasoning'' (Glenn Shafer, cited in~\cite{DBLP:books/daglib/0066829}).

This method allows us to learn probabilities (beliefs) from data (Sections~\ref{sec:learning} and~\ref{sec:experiments})  and thus ground their nature in the observation of accepted or rejected arguments.
Moreover, to study the properties of the joint probability distribution of arguments we can rely on the traditional inference tasks of PLP frameworks.
Interestingly,
\cite{nilsson2014understanding} remarks that considering the consequences and the explanations for beliefs is fundamental for critical thinking, and we can do precisely this by means of two typical PLP tasks: conditional queries and most probable explanation queries.
\paragraph{Conditionals.} We can perform conditional reasoning on the joint probability distribution by querying the probability of atoms given some evidence. We can thus analyze the consequences of beliefs by answering conditional queries such as: ``What are the consequences of accepting $a_1$?''. Note that $a_6$ is \emph{conditionally independent} of the other arguments  \emph{given} $a_1$, because when $a_1$ is known to be accepted (or rejected) the belief in $a_6$ depends only on the corresponding attack and bias.\newpage
\begin{example}\label{ex:evidence_arg}
    By querying the model from Example~\ref{ex:enccycle} with the addition of the evidence $arg(a_1)=true$, usually simply denoted with $arg(a_1)$, we infer the following conditional probabilities: $\mathbb{P}(arg(a_1)|arg(a_1))=1$, $\mathbb{P}(arg(a_2)|arg(a_1))=0.08$, $\mathbb{P}(arg(a_3)|arg(a_1))=0.36$, $\mathbb{P}(arg(a_4)|arg(a_1))=0.75$, \sloppy$\mathbb{P}(arg(a_5)|arg(a_1))=0.58$, $\mathbb{P}(arg(a_6)|arg(a_1))=0.28$. We can also condition the distribution on the relations, for instance by adding the evidence that accepting $arg(a_1)$ always excludes $arg(a_6)$. In this case, if we make explicit the belief in the attack with a probabilistic fact $\mathit{att}$ we obtain the same distribution except for  $\mathbb{P}(arg(a_6)|arg(a_1),\mathit{att})=0$.
\end{example}

The conditional independence assumption materializes also when we introduce a belief update to an argument which is conditionally independent of another, given the evidence for an argument that represents the connection of the two beliefs.
\begin{example}
    Consider again Example~\ref{ex:evidence_arg} with evidence $arg(a_1)=true$. If we add the evidence $arg(a_5)=false$, let $ev=(arg(a_1)=true\land arg(a_5)=false)$, then we obtain the following marginals: $\mathbb{P}(arg(a_1)|ev)=1$, $\mathbb{P}(arg(a_2)|ev)=0.08$, $\mathbb{P}(arg(a_3)|ev)=0.36$, $\mathbb{P}(arg(a_4)|ev)=0.62$, \sloppy$\mathbb{P}(arg(a_5)|ev)=0$, $\mathbb{P}(arg(a_6)|ev)=0.28$. The only beliefs affected by the new information about $a_5$ are those for $a_5$ and $a_4$, because $a_3$ is independent of $a_5$, $a_1$ is known, and $a_6$ and $a_2$ are \emph{conditionally independent given} $a_1$.
\end{example}

Conditional independence can also be exploited in argumentation to select arguments that provide new information with respect to an argument or claim. If these arguments' uncertainty can be eliminated by means of experiments or observations, the conditional independence assumption allows us to select only those strictly necessary to influence the belief in the argument of interest.

\begin{example}
    Bayesian statistical methods are suited for designing medical clinical trials to find an optimal design which minimizes the empirical results required and at the same time maximizes the information provided regarding the quality of the trial~\citep{berry2006bayesian}. Given an argumentative graph with respect to some treatment hypothesis, if some experimental results are unknown then they are associated with a probability (degree of belief). The uncertainty can be removed by collecting empirical evidence, but this comes at a cost of time and resources. In this case, the conditional independence of the claim with respect to some experiments given other empirical results can be exploited to select only those experiments that truly affect the belief in the claim.
\end{example}

In \smp{} it is also possible to condition the theory to be consistent, by providing the evidence that an atom is not interpreted as inconsistent.
This allows us to reason about how beliefs change if they are conditioned to be consistent with the logical structure.
\begin{example}
    Consider Example~\ref{ex:barber}: if we add the evidence $villager(bob) = \lnot inconsistent$ then we consider only the consistent possible worlds. Therefore, the probability of each possible world is normalized w.r.t. the probability mass of the consistent total choices. Hence, the probability that bob is both a villager and a barber becomes 0, because there are no models for consistent worlds where this is true. Each consistent possible world then increases its probability from $0.25$ to $\frac{1}{3}$. For instance $\mathbb{P}(villager(bob) | villager(bob)=\lnot inconsistent) = \frac{1}{3}.$
\end{example}

\paragraph{MPE.}\emph{Most probable explanation} (MPE) is a typical PLP task that allows us to address the other point of critical thinking, reasoning on explanations for beliefs. The MPE task is to compute the most likely possible world where some given evidence holds.

\begin{example}
    If we compute a MPE query for $arg(a_1)= true$, we are asking which set of beliefs compatible with accepting $a_1$ is the most probable. This means finding the most likely set of probabilistic facts representing a choice of biases and relations with the consequence of accepting $a_1$.
    In the MPE world the biases believed to support the corresponding argument are $\{bias(a_1).\, bias(a_2).\, bias(a_4).\, bias(a_5).\, bias(a_6).\}$. All relations are believed to hold, except for the attack of $a_4$ to $a_1$. $a_1$ is thus independent of $a_4$ and $a_5$ and its acceptance is supported by its bias and the non-deterministic choice with $a_2$ (following the evidence). These choices, that is, the MPE, have probability $0.0093$. Their model is $\{arg(a_1), \lnot arg(a_2), \lnot arg(a_3), arg(a_4), arg(a_5), \lnot arg(a_6)\}$ which is thus compatible with the evidence of accepting $arg(a_1)$. The two arguments attacked by $a_1$, $a_2$ and $a_6$, are rejected, along with $a_3$ because of the choice of not considering a bias towards it.
\end{example}

\subsection{Epistemic properties}
The joint probability distribution describes the subjective assignment of beliefs to arguments coherent with all possible belief choices of prior bias and relations, regardless of arbitrary thresholds. This means that the joint beliefs are globally coherent with the argumentative structure and therefore the marginal beliefs are also locally coherent with the parents and children's beliefs. The properties for epistemic probabilistic argumentation by~\cite{DBLP:conf/sum/PolbergHT17} can thus be used to evaluate the bias of the agent, which is indeed an arbitrary assignment of values, but in evaluating the joint probability distribution, having probability higher than 0.5 has no special meaning.
We clarify this by considering some of these properties and show that the marginal belief of the arguments defined by the joint distribution is compatible with the given argumentative structure even if they may not satisfy such properties.

\paragraph{Coherency.} \cite{DBLP:conf/sum/PolbergHT17} define coherency in epistemic probabilistic argumentation as the property that for all attacks $(a,b)$ believed with probability $>0.5$, $P_A(a)\leq 1-P_A(b)$.
In the joint probability distribution the marginal probability of arguments $\mathbb{P}(arg(\_))$ is coherent not just locally, considering pairwise marginals, but globally, where the structure of all relations and biases is taken into account.
\begin{example}
    Consider the case where two arguments $a$ and $b$ are such that $a$ attacks $b$ with probability $0.6$ and both arguments have bias (prior) 1. $a$ is not attacked so $P_A(a)=\mathbb{P}(arg(a))=1$. We can say that the initial assignment, the priors, is not coherent in the sense of~\cite{DBLP:conf/sum/PolbergHT17} because $P_A(a)=1\not\leq 1-1$. However, the joint probability distribution defines posterior marginals that are coherent in the sense of~\cite{nilsson2014understanding}: $\mathbb{P}(arg(b))=0.4$ because the uncertainty about the attack leaves room for possible worlds where the attack is not believed to hold (probability $1-0.6=0.4$) and thus it is possible to accept both $a$ and $b$. Here $1=\mathbb{P}(arg(a))\not\leq 1-\mathbb{P}(arg(b))=0.6$, but $\mathbb{P}(arg(a))$ should not be less or equal than 0.6 because it is independent of $b$, while accepting $b$ is influenced by accepting $a$ and its attack, and thus its prior bias, not the attacker, is inhibited to $0.6$.
\end{example}

\paragraph{Rational.} A rational (resp. strict) assignment resembles the conflict-free principle that if an attacker $a$ is accepted than the target $b$ should not: $P_R((a,b))>0.5$ and $P_A(a) > 0.5$ implies $P_A(b)\leq 0.5$ (resp. $P_A(b)<0.5$). In our framework, when choosing a set of prior beliefs (probabilistic facts) the conflict-freeness of the accepted (true) arguments is always guaranteed by the logic rules.
This does not exclude the possibility that both $a$ and $b$ in the joint probability distribution are believed with probability $>0.5$. In fact, when an argument has high bias, e.g. we think it is almost certain, only a combination of strong attackers with strong relations can lower its inferred probability under 0.5.
\begin{example}
    For example, a prior probability of $0.6$ of believing in $a$ and $0.6$ in the attack, yields an inhibition by a factor of $0.36$, which is not guaranteed to lower the marginal probability of $b$ under $0.5$. In $36\%$ of the possible worlds $b$ is rejected because of the presence of $a$ and the attack, but if $P_A(b)=1$ then $\mathbb{P}(arg(b))=0.64$, together with $\mathbb{P}(arg(a))=0.6$ and $P_R((a,b))=0.6$. $\mathbb{P}(arg(b))\leq 0.5$ only when $\mathbb{P}(a)\cdot P_R((a,b))\geq 0.5$ (in the case of inhibition from a single attack). This does not mean that the joint probability distribution is not rational in the sense that it defines a set of accepted arguments not conflict-free. The joint probability distribution reflects the relative strength of the arguments and simply defines a (coherent) probability $\geq 0.5$ to be in a possible world where either $a$ or $b$ are accepted, but there is no possible world where the attack is believed and both are accepted.
\end{example}

\paragraph{Protective.} A (restricted) protective assignment is such that $P_R((a,b))\geq 0.5$ (resp. $P_R((a,b))>0.5$) and $P_A(b) > 0.5$ implies $P_A(a)\leq 0.5$. Similarly to the rational property, in epistemic probabilistic argumentation this is the principle of rejecting the attacker if the target is accepted. Again, this principle in the joint belief distribution is not enforced by means of the probability (belief) values, but by means of the logical structure of the attack relation.
\begin{example}
    Our previous example  also describes a belief distribution that is not protective: while $\mathbb{P}(b)=0.64$ and $P_R((a,b))=0.6$, $\mathbb{P}(a)$ remains equal to its original prior $P_A(a)=0.6$, thus $\mathbb{P}(arg(a))>0.5$. This does not mean that both arguments are accepted \emph{because} their probability is higher than 0.5, but it means that a joint belief induced by the argumentative structure coherent with the prior bias assigns a subjective marginal belief $\mathbb{P}(arg(a))$ and $\mathbb{P}(arg(b))$ higher than 0.5 to the arguments. The difference in strength is still coherent with the respective biases and belief in the attack relation and probability axioms.
\end{example}

Similar arguments can be constructed for the other properties from~\cite{DBLP:conf/sum/PolbergHT17}: past epistemic argumentation approaches consider the probabilities of the argument graph as the \emph{marginal} beliefs of the agent and check if those are (locally) rational, coherent, \dots Our approach uses this arbitrary assignment as the \emph{prior} set of beliefs, and the argumentative structure, to infer a joint belief distribution that is compatible with the initial arbitrary assignment according to the probability postulates and the logical relations of the arguments.
The joint belief (probability) distribution thus represents how the prior beliefs (biases) interact with each other to define \emph{marginal and conditional} beliefs that are coherent with the given priors  and argumentative (logic) structure.

\section{Inference and learning}\label{sec:implementation}

In this section we describe the inference and learning algorithms for \smp{} semantics. We consider the inference and learning algorithms implemented in ProbLog2 and adapt them to the semantics defined in Section~\ref{sec:semantics}. This allows us to study the advantages and shortcomings of traditional PLP inference and learning techniques when a more general semantics is concerned, and indirectly examine their application to argumentation problems.
\subsection{Inference task}\label{sec:inference}
\begin{figure}[b!]
    \centering

    \begin{tikzpicture}[->,>=stealth',shorten >=1pt,auto,node distance=2.8cm,
            semithick, every text node part/.style={align=center}]
        \tikzstyle{every state}=[fill=white,shape=rectangle]

        \node[state]         (Z)                    {Program $\mathcal{L}$};
        \node[state]         (A) [right=1.8cm of Z] {Ground $\mathcal{L}$};
        \node[state]         (B) [right=1.5cm of A]       {CNF};
        \node[state]         (C) [right=1.8cm of B] {Logic Circuit};
        \node[state]         (D) [below=1cm of C]  {Artithmetic Circuit};
        \node[]         (E) [left=1cm of D]  {Evaluation (WMC)};

        \path (Z) edge []  node {Grounding} (A);
        \path (A) edge []  node {Cycle\\breaking} (B);
        \path (B) edge []  node {Knowledge \\compilation} (C);
        \path (C) edge []  node {logic to arithmetic \\operators} (D);
        \path (E) edge [dashed]  node {} (D);
    \end{tikzpicture}
    \caption{ProbLog2 inference schema}
    \label{fig:problog_pipeline}

\end{figure}
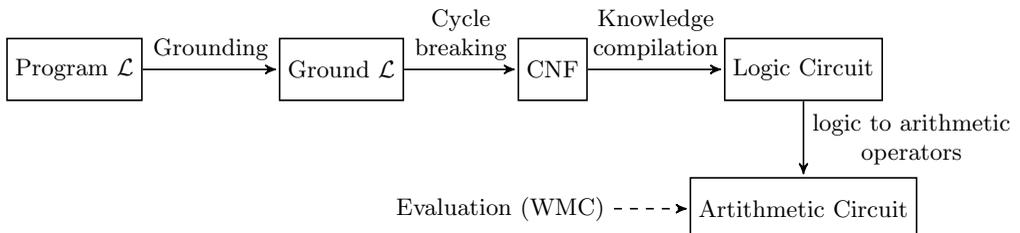

The state-of-the-art inference technique for probabilistic logic programs is reducing the inference task to a weighted model counting problem (cfr. Section~\ref{sec:background}), therefore we consider its applicability in the context of our semantics.
In ProbLog2 inference is implemented by means of a transformation of the ground program to Conjunctive Normal Form, which is the input for a knowledge compiler (under well-founded semantics) that returns the logical circuit (e.g. a d-DNNF) such that on the corresponding arithmetic circuit the WMC task can be solved in polynomial time (Figure~\ref{fig:problog_pipeline}). In the rest of the section we describe the adaptations required to this inference technique required for {\smp}'s semantics.

\begin{figure}[t]
    \centering

    \begin{tikzpicture}[->,>=stealth',shorten >=1pt,auto,node distance=2.8cm,
            semithick, every text node part/.style={align=center}]
        \tikzstyle{every state}=[fill=white,shape=rectangle]

        \node[state]         (A) [right=1cm of A]                  {Ground $\mathcal{L}$};
        \node[state]         (B) [right=2.2cm of A]       {\textsc{dsharp} input\\format\textsuperscript{*}};
        \node[state]         (C) [right=1.8cm of B] {Logic Circuit};
        \node[state]         (D) [below=1cm of C]  {Artithmetic Circuit};
        \node[]         (E) [right=1cm of C]  {(3) Enumeration };
        \node[]         (F) [left=1cm of D]  {(4) Evaluation ($\mathit{\widehat{WMC}}$)};
        \node[state]         (G) [above=1cm of A]                  {Program $\mathcal{L}$};

        \path (G) edge []  node {(1) Grounding} (A);
        \path (A) edge []  node {Preprocessing} (B);
        \path (B) edge []  node {(2) Knowledge \\compilation} (C);
        \path (C) edge []  node {logic to arithmetic \\operators} (D);
        \path (E) edge [dashed]  node {} (C);
        \path (F) edge [dashed]  node {} (D);
    \end{tikzpicture}
    \caption{{\smp} inference schema. (*) denotes a different version of \textsc{dsharp} from ProbLog2, specific for stable model counting.}
    \label{fig:smproblog_pipeline}

\end{figure}
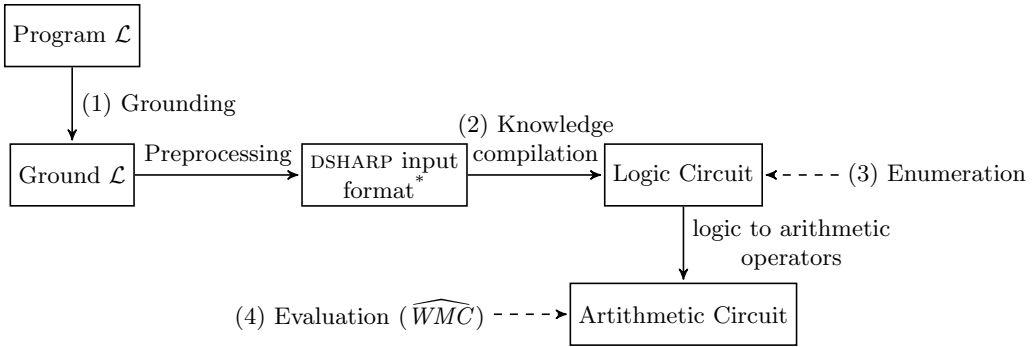

The normalization of the weight of the models entails considering a variant of the $\mathit{WMC}$ problem where models are weighted with the corresponding normalization constant. We denote such constant with $\hat{w}(M_\omega)$, where $M_\omega$ is a model for a total choice $\omega$, hence $\hat{w}(M_\omega)=\frac{1}{|\mathit{MOD}(\mathcal{L},\omega)|}$. The $\mathit{WMC}$ problem ($\mathit{WMC}_{\mathcal{L}}(\varphi)= \sum_{M\in \mathit{MOD}(\mathcal{L}),M\models \varphi}\prod_{l\in M}w(l)$) thus becomes: \[\widehat{\mathit{WMC}}_{\mathcal{L}}(\varphi)= \sum_{M\in \mathit{MOD}(\mathcal{L}),M\models \varphi}\hat{w}(M)\cdot\prod_{l\in M}w(l).\]
This formulation of the problem requires us to determine the weights $\hat{w}(M_\omega)$, which, contrary to the weights of the literals, are not known from $\mathcal{L}$. In order to determine the weighting function $\hat{w}$ we thus need to solve an additional counting problem, namely the task of counting how many models agree on a subset of atoms in $\mathcal{L}$, that is, the probabilistic facts.
We now describe the reduction pipeline in \smp{} to a $\mathit{\widehat{WMC}}$ problem. The reduction can be broadly divided into four components, represented in Figure~\ref{fig:smproblog_pipeline}: (1) grounding the program $\mathcal{L}$, (2) compiling the logic theory $\mathcal{L}$ into a d-DNNF, (3) deriving a circuit for retrieving the number of models corresponding to a total choice and an arithmetic circuit for the $\mathit{WMC}_{\mathcal{L}}$ problem (4) the evaluation of the two circuits to define the solution of the $\mathit{\widehat{WMC}_{\mathcal{L}}}$ problem (Algorithm~\ref{algo:evaluate}).

\paragraph{1) Grounding.} In \smp{} we adopt a standard bottom-up grounding technique. In ProbLog2, the detection of cycles through negations during the grounding step is used as a sufficient condition to reject invalid programs. On the contrary, we use the absence of cycles through negations as a condition to reduce to ProbLog's evaluation algorithm, where \smp's semantics is equivalent, as discussed in Section~\ref{sec:related_prob}. In fact, when this is the case, we skip the enumeration step (Step 3) because $\hat{w}(M)=1$ for all stable (and well-founded) models $M$. Therefore, the
$\widehat{\mathit{WMC}}_{\mathcal{L}}$ problem becomes equivalent to a $\mathit{WMC}_{\mathcal{L}}$ problem.

\paragraph{2) Compilation.}

\begin{figure}[t]

    \begin{minipage}[t]{0.69\textwidth}
        \centering
        \vspace{0pt}
        \includegraphics[width=\textwidth]{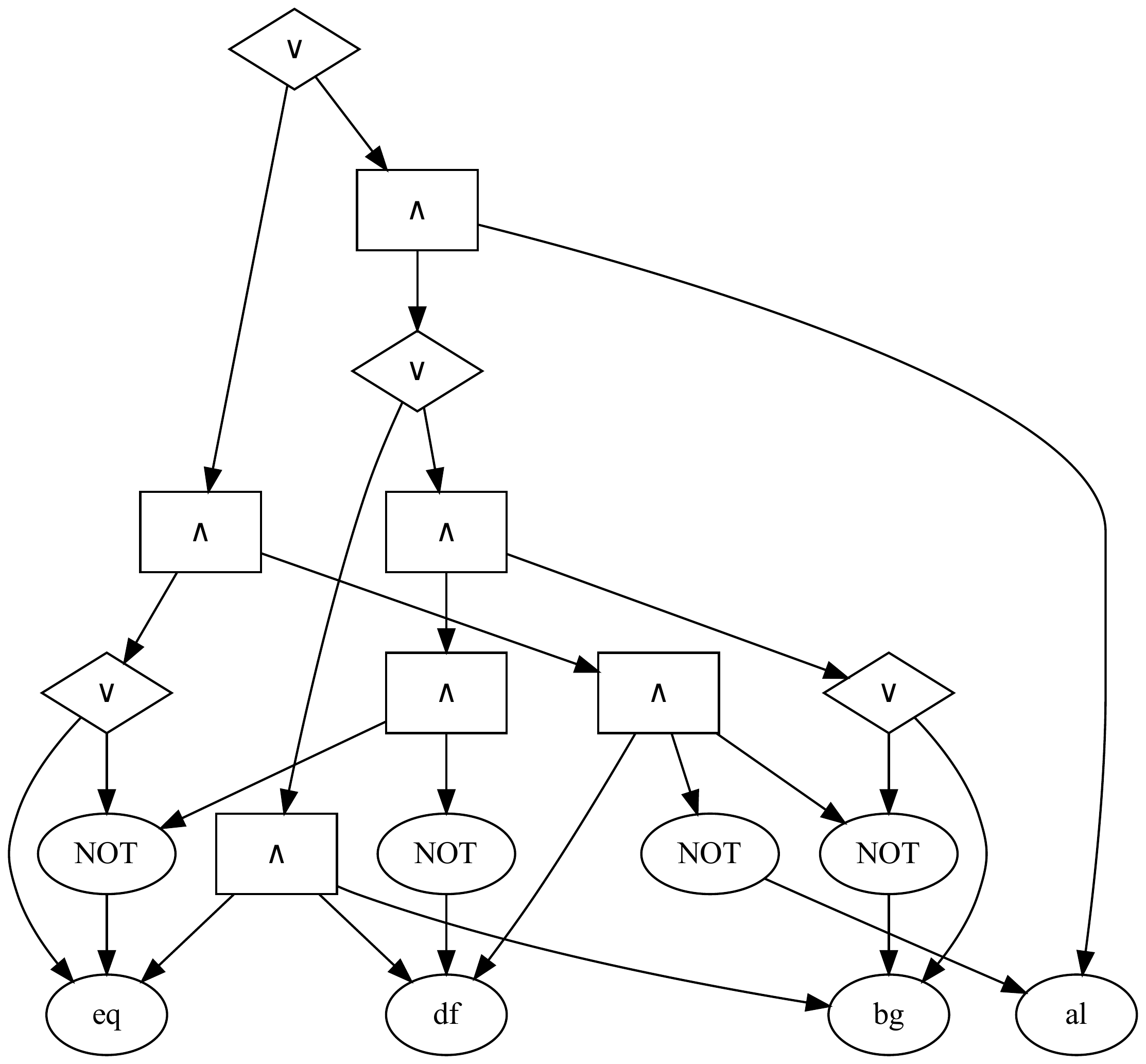}
    \end{minipage}
    \hfill
    \begin{minipage}[t]{0.3\textwidth}
        \centering
        \vspace{0pt}
        \hspace{-2.5cm}
        \setlength{\fboxsep}{0pt}\framebox{\includegraphics[width=1.5\textwidth]{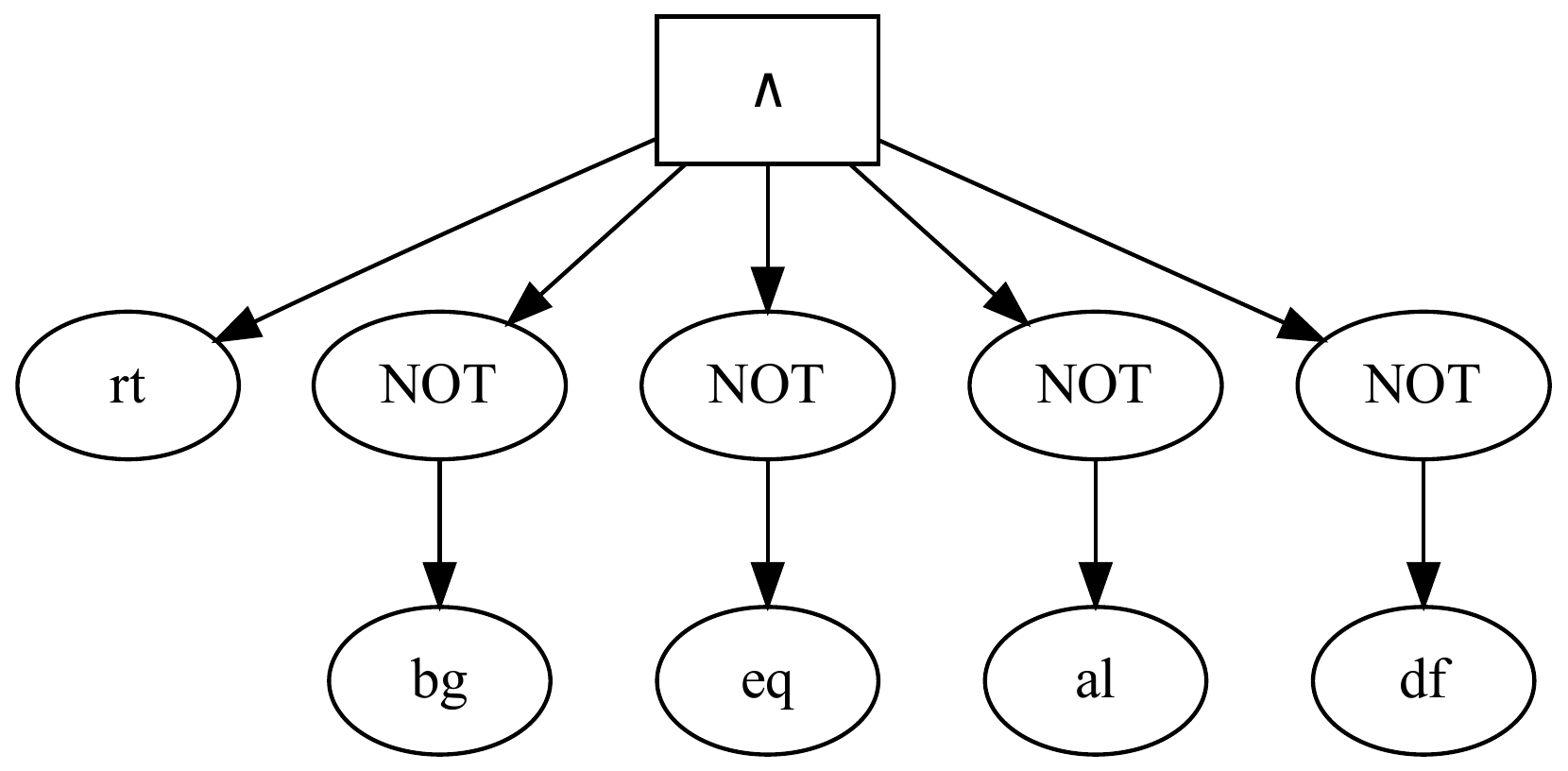}}
    \end{minipage}
    \caption{Compilation step of Example~\ref{ex:zero} (top right) and Example~\ref{ex:multiple} (left). Models for both theories are compactly encoded as d-DNNF logical circuits. $bg=\bg$, $eq=\eq$, $al=\al$, $df=\df$, $rt=\rt$.}
    \label{fig:kc}
\end{figure}

The goal of the compilation step is to transform the ground propositional theory in a representation that allows us to answer efficiently queries about the probability of atoms. Knowledge compilation produces a logical circuit describing the possible worlds' models of the input propositional theory. Therefore, this step regards exclusively the logical part of the program, while the following inference steps concern the computation of probabilities.
As in Problog2, the knowledge compiler is a black box with respect to the probabilistic inference system.
Therefore, in \smp{} we are only concerned with transforming the propositional ground theory to the suitable input format. The input format and the implementation details of the knowledge compiler in \smp{} can be found in~\citeN{DBLP:conf/aaai/AzizCMS15}, which extends the \textsc{dsharp} compiler~\citep{DBLP:conf/ai/MuiseMBH12} with stable model semantics. In ProbLog2 the input format is a theory in Conjunctive Normal Form (CNF), \citeN{DBLP:conf/aaai/AzizCMS15} replace it with an encoding in CNF format of the rules, variables and the corresponding strongly connected components. The output is a logical circuit, a d-DNNF, describing the \emph{stable} models of the input propositional theory.
The circuit therefore expresses only the models corresponding to the consistent total choices.
This representation, with a standard transformation in an arithmetic circuit, allows efficient stable model counting, that is, solving the $\mathit{WMC}$ problem.
Figure~\ref{fig:kc} shows the output of the knowledge compilers from the representations of Example~\ref{ex:zero} and~\ref{ex:multiple}.
Contrary to ProbLog2, however, the transformation of the logical circuit into an arithmetic circuit is not sufficient to solve the $\widehat{\mathit{WMC}}$ problem. This because the arithmetic circuit allows us to efficiently compute the solution to a $\mathit{WMC}$ problem, but the normalization constants $\hat{w}$ are still unknown. Unless the weighting is determined by the absence of cycles through negations, as defined in the previous paragraph, the enumeration step is required to associate each model to the corresponding normalization constant.

\paragraph{3) Enumeration.}
\begin{figure}[t]

    \begin{minipage}[t]{0.70\textwidth}
        \centering
        \vspace{0pt}
        \includegraphics[width=\textwidth]{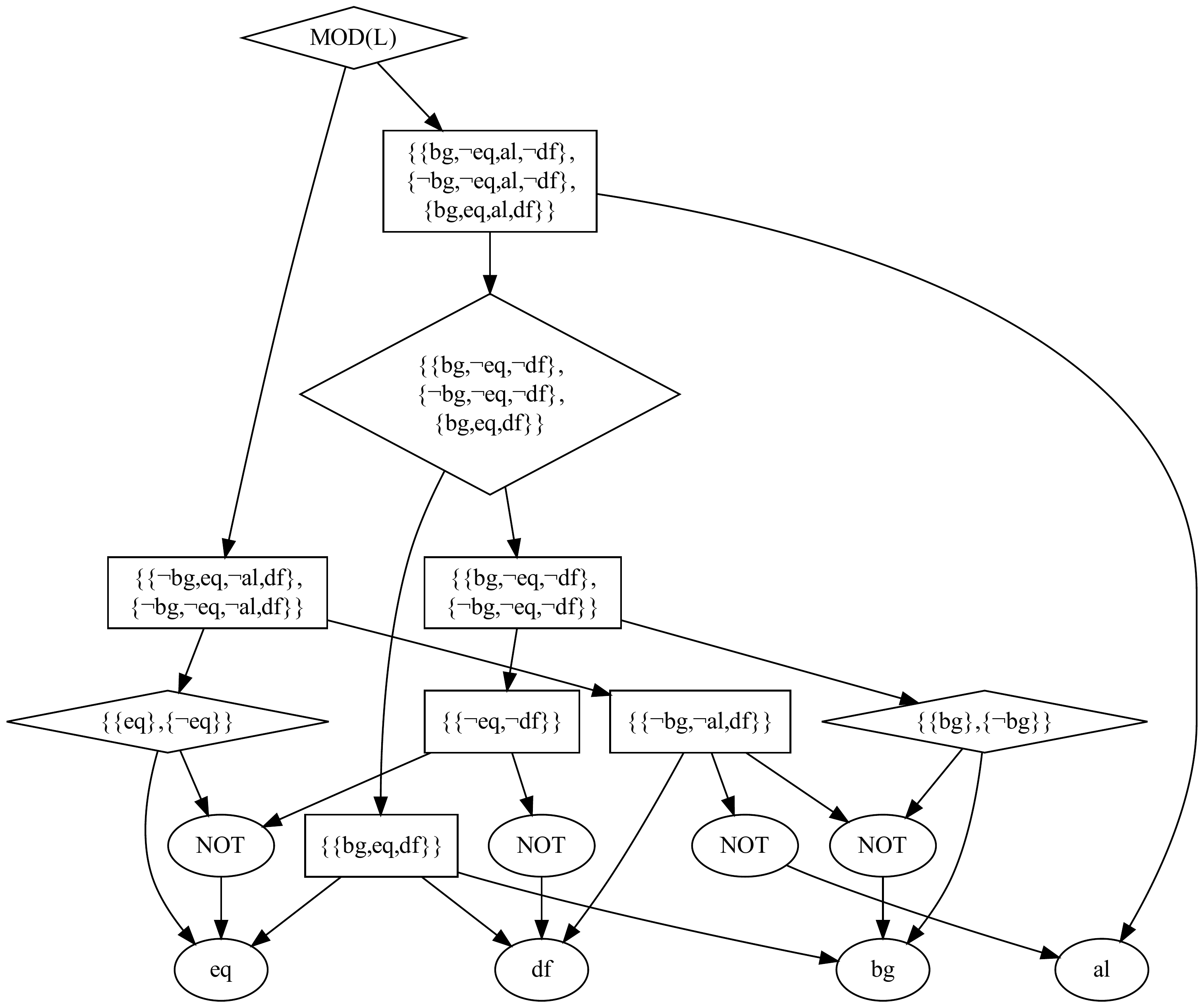}
    \end{minipage}
    \hfill
    \begin{minipage}[t]{0.29\textwidth}
        \centering
        \vspace{0pt}
        \hspace{-2.75cm}
        \setlength{\fboxsep}{0pt}\framebox{\includegraphics[width=1.5\textwidth]{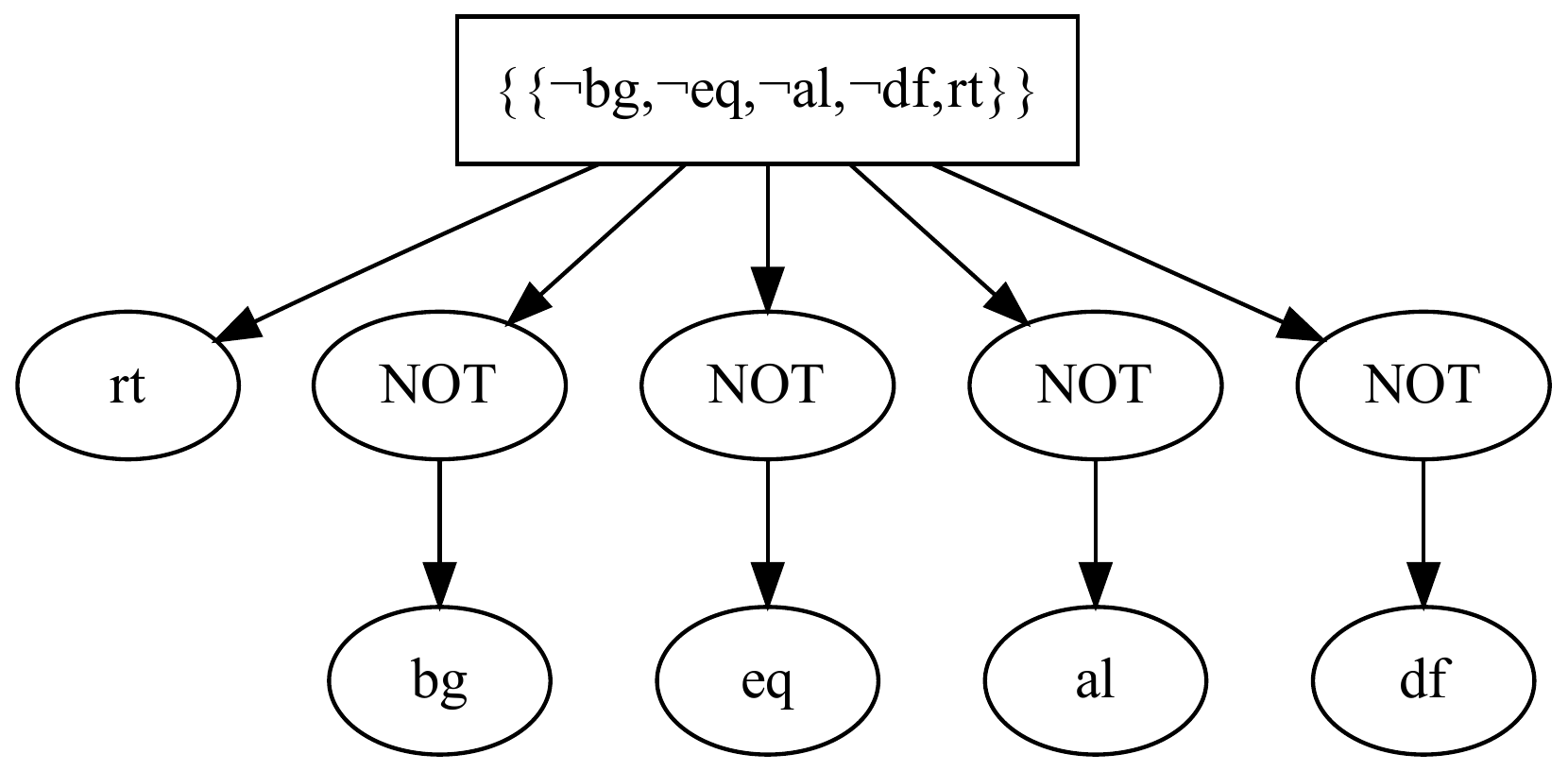}}
    \end{minipage}
    \caption{Enumeration step of Example~\ref{ex:zero} (top right) and Example~\ref{ex:multiple} models (left). Each node contains the corresponding (partial) models. Diamonds are union nodes, squares are Cartesian products. $M$ is the complete list of the 5 models of Example~\ref{ex:multiple}.}
    \label{fig:enum}
\end{figure}
The purpose of the enumeration step is to count the number of stable models corresponding to each (consistent) total choice, which defines $\hat{w}$. Unfortunately, this operation cannot be performed efficiently on the representation obtained from the knowledge compiler, because models corresponding to the same total choice may be represented by different nodes and subtrees. Therefore, on this kind of representation we are forced to traverse the circuit in order to retrieve the list of models and the corresponding total choices.

We thus derive from the d-DNNF a circuit where each leaf for a literal $l$ is replaced by a set of (partial) models $\{\{l\}\}$, disjunctions are replaced by the union of the children and conjunctions correspond to the Cartesian product of the children (Figure~\ref{fig:enum}).
Traversing bottom-up such circuit returns the list of models, from which we build a map $\#:\Omega_\mathcal{L}\rightarrow\mathbb{N}$ from total choices to the corresponding number of models, i.e. $\#(\omega)=|\mathit{MOD}(\mathcal{L},\omega)|$. The reciprocal thus defines the normalization constants $\hat{w}$. For instance, in Example~\ref{ex:multiple} we obtain $\#(\omega_1)=\#(\omega_2)=\#(\omega_3)=1$, $\#(\omega_4)=2$, hence $\hat{w}(\{\bg,\al,\eq,\rt\})=\hat{w}(\{\bg,\al\})=\hat{w}(\{\eq,\mathit{\df}\})=1$, and $\hat{w}(\{\al\})=\hat{w}(\{\mathit{\df}\})=\frac{1}{2}$.
Note that the total choices that do not appear in the enumeration of models are those that have zero models and thus are inconsistent. Because probabilistic facts are disjoint from the rules' heads, for each model described by the circuit the truth value of the literals corresponding to probabilistic facts defines the corresponding total choice (they can be true iff chosen by $\omega$).

The circuit traversal is a computationally expensive step since it entails enumerating all possible models. Contrary to sums (logical ORs) and products (logical ANDs), the Cartesian product is not a linear-time operation. Therefore, the computational cost of the internal nodes raises with the length of the lists of partial models corresponding to the children. For this reason, the study of novel knowledge compilation techniques that produce a representation where the normalization constants can be obtained from linear time operations is an interesting direction for improvement.

This is a novel problem because \smp{} semantics applies a different normalization constant per total choice ($2^n$ with $n$ facts), while in probabilistic ASP the normalization constant is the same for all models, that is, the weight of all possible worlds. This can be obtained with a single evaluation of the weight of the root of the circuit. Therefore, obtaining the normalization constants from a circuit where node-level operations are computable in linear time would remove this bottleneck, which is introduced because we consider a fundamentally different inference task.

\begin{figure}[t]
    \begin{minipage}[t]{0.8\textwidth}
        \centering
        \vspace{0pt}
        \includegraphics[width=\textwidth]{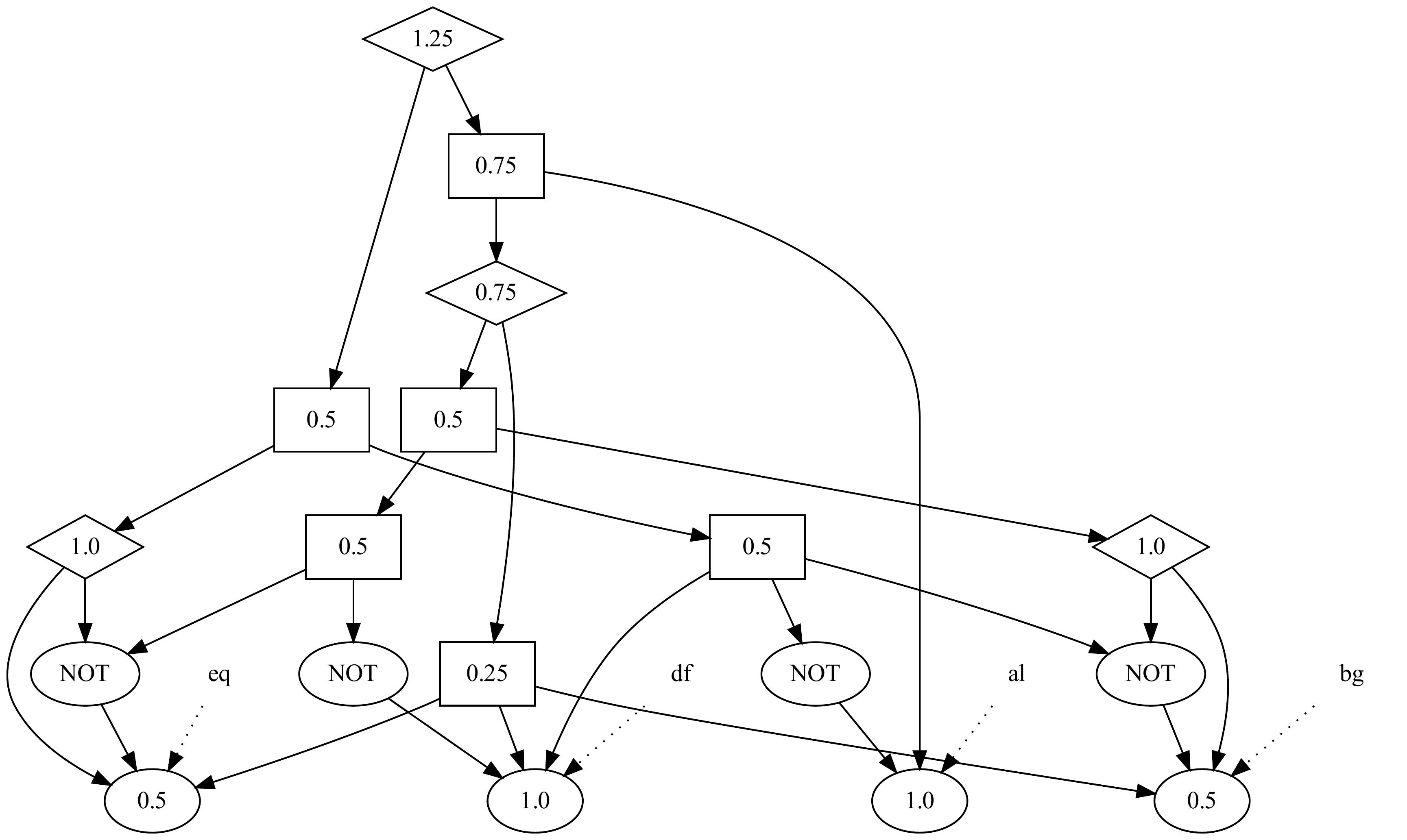}
    \end{minipage}
    \hfill
    \begin{minipage}[t]{0.19\textwidth}
        \centering
        \vspace{0pt}
        \hspace{-4cm}
        \fbox{\includegraphics[width=2.1\textwidth, trim=0cm 0.7cm 3cm 0cm]{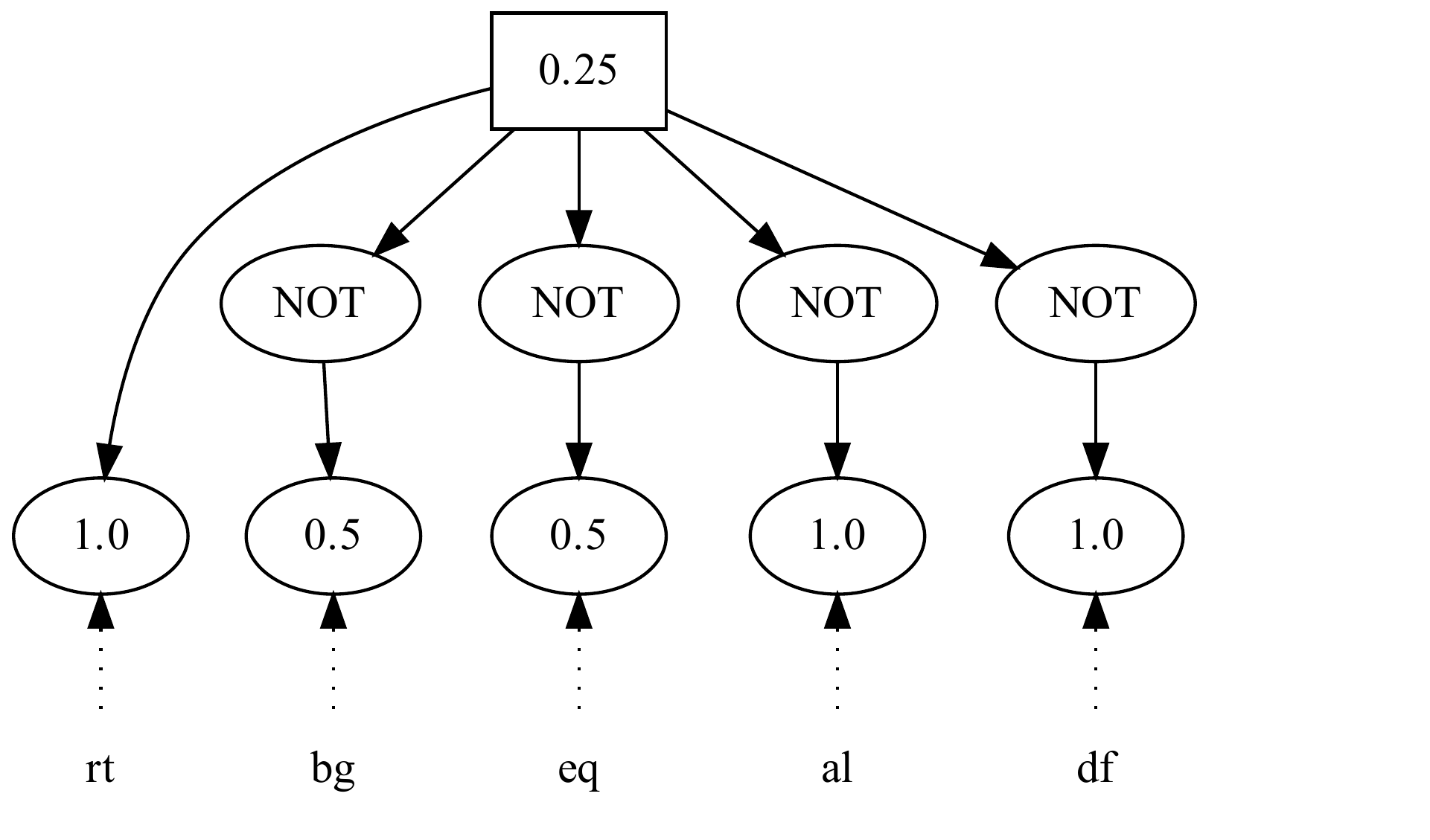}}
    \end{minipage}
    \caption{$\mathit{WMC}$ of Example~\ref{ex:zero} (top right) and Example~\ref{ex:multiple} (left) . Each node contains the corresponding weight, logical variables do not influence the weights. Diamonds are sum nodes, squares are product nodes.}
    \label{fig:WMC}
\end{figure}%

\paragraph{4) Evaluation.} As we previously mentioned, to evaluate $P(q\,|\,E=e)$, with $\varphi=q\land E=e$, on a program $\mathcal{L}$ we solve the problem $\widehat{\mathit{WMC}}_{\mathcal{L}}(\varphi)= \sum_{M\in \mathit{MOD}(\mathcal{L}),M\models \varphi}\hat{w}(M)\prod_{l\in M}w(l)$.
The weights of the leaves in the arithmetic circuit are instantiated according to the query and the given evidence (Section~\ref{sec:background}).
\citeN{DBLP:conf/uai/FierensBTGR11} show that the arithmetic circuit derived from the d-DNNF computes $\mathit{WMC}_{\mathcal{L}}(\varphi)= \sum_{M\in \mathit{MOD}(\mathcal{L}),M\models \varphi}\prod_{l\in M}w(l)$. Under a one-to-one correspondence between models and total choices, given $\omega\in\Omega_\mathcal{L}$ and  its model $M_{\omega}$, $\mathit{WMC}(M_\omega) = \mathit{WMC}(\omega)$. Therefore, if the enumeration step is skipped, i.e. $\hat{w}(M_{\omega})=1$ for all stable (and well-founded) models $M_{\omega}$, then $\mathit{WMC}(M_\omega)  = \mathit{WMC}(\omega) = \widehat{\mathit{WMC}}(M_\omega)$.

On the contrary, if there are multiple models for some total choices, then $\mathit{WMC}(M_\omega)  = \mathit{WMC}(\omega) \neq \widehat{\mathit{WMC}}(M_\omega)$, because $\mathit{WMC}(M_{\omega})$ overcounts the normalized weights of the models corresponding to the same total choice.
Figure~\ref{fig:WMC} shows the weight calculation of the root node of the circuit: the root corresponds to all consistent total choices (e.g. Figure~\ref{fig:enum}). Example~\ref{ex:multiple} shows that the original algorithm overcounts the weight when multiple stable models correspond to the same total choice. This because, their weight is not normalized (cfr. Example~\ref{ex:global} for the explanation of the 1.25 weight). This raises the question as to how the evaluation algorithm has to account for the normalization constants for \smp{} semantics.

As Figure~\ref{fig:enum} shows, multiple models can correspond to the same node, therefore it is not possible to associate a normalization constant to a single node. Moreover, the aggregation of the weights in one value per node, as in Figure~\ref{fig:WMC}, is no longer possible, because two partial models (e.g. $\{\{bg\},\{\lnot bg\}\}$) may be conjoined (multiplied) with partial models (weights) that require different normalization constants. Once the weights are aggregated it is not possible to normalize the different components of the weight with different constants.
We now describe two evaluation strategies to normalize weights in the evaluation step.

The first is essentially a repetition of the enumeration step: we traverse again the circuit bottom-up and along with each set of atoms we carry the corresponding weight: once a total choice is defined, the weights of each partial model are multiplied and normalized w.r.t. the corresponding number of models obtained from the enumeration step. The drawback of this strategy is that the complexity of the enumeration step is repeated at each query.

The second method is to exploit the efficiency of the $\mathit{WMC}$ task on the circuit and compute first the unnormalized weight of the root and then correct it (Algorithm~\ref{algo:evaluate}).
For each total choice $\omega$ s.t. $n=\#(\omega)>1$ and $\omega\models \varphi$, the unnormalized evaluation of the circuit overcounts the weight of its models, because they are not multiplied by $\hat{w}(M_\omega)=\frac{1}{n}$.
Remember that the unnormalized weight of a model is the probability of the corresponding total choice: $\mathit{WMC}(M_\omega) = \mathit{WMC}(\omega)=P(\omega)$, therefore the unnormalized weight is such that \[\mathit{WMC}(M_\omega)= \hat{w}(M_\omega)\cdot \mathit{WMC}(\omega) + (1-\hat{w}(M_{\omega}))\cdot \mathit{WMC}(\omega).\]
Since $\hat{w}(M_\omega)=\frac{1}{n}$ and $\mathit{WMC}(\omega)=P(\omega)$, each weight the of models $M_\omega\in \mathit{MOD}(\mathcal{L},\omega)$ is overcounted by \[(1-\hat{w}(M_{\omega}))\cdot\mathit{WMC}(\omega)=\frac{P(\omega)\cdot(n-1)}{n}.\]
Removing such weight from the unnormalized one, gives the probability defined by \smp{}'s semantics:
\[\widehat{\mathit{WMC}}(M_{\omega})=\mathit{WMC}(M_{\omega})-\frac{P(\omega)\cdot(n-1)}{n}=P(\omega)-\frac{P(\omega)\cdot(n-1)}{n}=\frac{P(\omega)}{n} = \hat{P}(M_{\omega}).\]

The overhead of this method per query lies in the iteration over the total choices with multiple models to retrieve and apply the different normalization constants. On the other hand, there is no additional cost in the operations performed at a node level while traversing the circuit, in contrast to the Cartesian product operations in the first method.
Algorithm~\ref{algo:evaluate} describes the correction of the weight for any formula $\varphi$, where only the models and total choices compatible with $\varphi$ contribute to $\mathit{WMC}(\varphi)$ with unnormalized weights.
In Section~\ref{sec:experiments} we show experimentally that exploiting the efficiency of the $\mathit{WMC}$ task results in the evaluation step being significantly faster than the enumeration step, whose complexity reflects on the first method.

Since the circuit describes only the stable models for consistent total choices, in the case of inconsistent programs we derive the probability of an inconsistency by removing from 1 the weight (probability) of the consistent total choices, as described in Section~\ref{sec:semantics}.

\begin{algorithm}
    \caption{Evaluation step: $\mathbb{P}(q | E=e)$}\label{alg:cap}
    \begin{algorithmic}
        \Require $\varphi = q\land E=e$
        \Ensure $P(q\,|\,E=e)$
        \State $w \gets \mathit{WMC}(\varphi)$
        \For{$\omega$ \textbf{in} $\Omega_\mathcal{L}$ \textbf{s.t.} $n=\#(\omega)>1,\omega\models E=e$ }
        \For{$M_{\omega}\in \mathit{MOD}(\mathcal{L},\omega)$ \textbf{s.t.} $ M_{\omega}\models \varphi$}
        \State $w = w-\frac{P(\omega)\cdot(n-1)}{n}$
        \EndFor
        \EndFor
        \State \Return $w$
    \end{algorithmic}
    \caption{Evaluation step schema}
    \label{algo:evaluate}
\end{algorithm}

\subsection{Learning task} \label{sec:learning}

In this section we show that the expectation-maximization learning algorithm implemented in ProbLog2~\citep{DBLP:conf/pkdd/GutmannTR11} is correct also for consistent \smp{} programs. With inconsistent programs, in fact, learning is not possible in a traditional EM setting, because we cannot associate a selection of probabilistic facts to an interpretation of inconsistent atoms. If one of the atoms is inconsistent then all other atoms are inconsistent. Therefore, we can estimate $\mathbb{P}(\mathcal{L}\models\bot)$ by counting the proportion of inconsistent observations, but we cannot learn how that estimate is distributed across the total choices causing an inconsistency. This information is essential, in this learning setting, to estimate the probability that an atom is chosen.~\citep{DBLP:journals/ml/Cussens01} extends the EM framework in \emph{stochastic logic programs}~\citep{muggleton1996stochastic} to account for failed derivation paths. In future work, investigating the applicability of his  \emph{failure adjusted maximization} algorithm to inconsistent probabilistic logic programs might offer a solution to overcome the limits of a traditional EM framework on inconsistent \smp{} programs.
On the other hand, we show that in the case of consistent \smp{} programs we can learn the probabilities of the program in both total and partial observability.

\paragraph{Total observability.}
The correctness under total observability is guaranteed by the fact that probabilistic facts are independent.
In fact, in total observability each interpretation $I_m\in \mathbb{I}$ observes the truth value of each atom and probabilistic fact of $\mathcal{L}$. This case reduces to counting the number of true occurrences of each probabilistic fact in the interpretations $\mathbb{I}$.
We clarify this by analyzing Equation~\ref{eq:learn_total} from \cite{DBLP:conf/pkdd/GutmannTR11} which is also implemented in \smp{}.
Let $\hat{p_n}$ be the estimate for $p_n::f_n$. Let  $\theta_{n,j}^m$ be the j-th possible grounding substitution for $f_n$ in the interpretation $I_m$. Interpretations $I_j$ thus represent a possible combination of ground substitutions for the probabilistic facts. Let $J^m_n$ be the number of such substitutions, and $Z_n=\sum^M_{m=1} J_n^m$ is the total number of ground instances of $f_n$ in all training examples:
\begin{equation}\label{eq:learn_total}
    \hat{p_n} = \frac{1}{Z_n} \sum^M_{m=1} \sum^{J^m_n}_{j=1} \delta_{n,j}^m \hskip0.05\textwidth \text{where }  \hskip0.05\textwidth\delta^m_{n,j}=
    \begin{cases}
         & 1 \text{ if } f_n\theta^m_{n,j}\in I_m; \\
         & 0 \text{ else}.
    \end{cases}
\end{equation}
The outermost sum ranges over all observed interpretations and the innermost sum counts each different observation of a ground instance of $f_n$. Each number of ground substitutions observed is then divided by the total number of ground substitutions. The estimate $ \hat{p_n}$ is thus defined as the proportion of ground substitutions observed over all interpretations observed.
This is correct for \smp{} programs because the observation of a probabilistic fact remains independent of the rest of the program for each interpretation.


\paragraph{Partial observability.}
The correctness under partial observability is guaranteed by the fact that the parameter updates rely on probabilistic inference, which is performed under \smp{}'s semantics and thus returns correct probability estimates.
In the partially observable case, where each interpretation $I_m$ observes the truth value of a subset of $\mathcal{L}$, the parameter update iteration relies on probabilistic inference to update the likelihood of a fact given $I_m$.
At each iteration $k$ the parameters from the previous iteration $k-1$ are used in $\mathcal{L}$ to compute the conditional expectation of the parameter given the interpretations, until convergence. The number of observations $\delta^m_{n,j}$ in (\ref{eq:learn_total}) is replaced by the conditional expectation under the current model $\mathbb{E}[\delta^m_{n,j}|I_m]$:
\begin{equation}\label{eq:learning}
    \hat{p_n} = \frac{1}{Z_n} \sum^M_{m=1} \sum^{J^m_n}_{j=1} \mathbb{E}[\delta^m_{n,j}|I_m].
\end{equation}


To compute the conditional expectation values we query the model for $\mathbb{P}(f_n)$ at each step under the current parameters estimate. This means that the normalization w.r.t. multiple stable models is incorporated in the conditional probabilities computed by the inference task, therefore also the parameter estimate is normalized w.r.t. the different stable models that may correspond to a given partial observation.  Since the algorithmic procedure for learning  in \smp{} remains the same as ProbLog2, except from the inference step, Example~\ref{ex:base_learn} is a valid example also for learing from a partial interpretation in \smp{}.

\section{Experiments}\label{sec:experiments}

The goal of our experiments is to establish the feasibility of our approach, and to answer the following questions: \begin{itemize}
    \item[(Q1)] What is the computational cost of the different steps in the pipeline for \smp{} semantics on the inference task?
    \item[(Q2)] Given a set of observations of accepted arguments generated from a known ProbLog program, can we learn the degrees of belief of the agent that modelled the original program?
    \item[(Q3)] How does the inference computational cost reflect on learning?
\end{itemize}

\paragraph{Q1. Inference.}

\begin{figure}[b]
    \centering

    \begin{minipage}[c]{0.45\textwidth}
        \vspace{0pt}
        \begin{tikzpicture}
            \pgfplotsset{%
                width=\textwidth,
                height=\textwidth
            }
            \begin{axis}[
                    ybar stacked,
                    bar width=15pt,
                    ymajorgrids,
                    grid style=dashed,
                    enlargelimits=0.15,
                    legend style={at={(0.5,1.15)},
                            anchor=north,legend columns=-1},
                    ylabel={time (s)},
                    symbolic x coords={$t_1$,$t_2$,$t_3$,$t_4$,$t_5$,$t_6$},
                    xtick=data,
                    x tick label style={rotate=45,anchor=east},
                    ytick={0,10,20,30,40}
                ]
                \addplot+[ybar] plot coordinates {($t_1$,0.006) ($t_2$,0.006)
                        ($t_3$,0.007) ($t_4$,0.007) ($t_5$,0.012) ($t_6$,0.019) };
                \addplot+[ybar] plot coordinates {($t_1$,0.008) ($t_2$,0.116)
                        ($t_3$,2.158) ($t_4$,5.337) ($t_5$,14.449) ($t_6$,32.541) };
                \addplot+[ybar] plot coordinates {($t_1$,0.110) ($t_2$,0.603)
                        ($t_3$,1.546) ($t_4$,2.158) ($t_5$,2.982) ($t_6$,7.460) };

                \legend{\strut (2) compile, \strut (3) enumerate, \strut (4) evaluate}
            \end{axis}
        \end{tikzpicture}
    \end{minipage}
    \quad
    \begin{minipage}[c]{0.5\textwidth}
        \vspace{0pt}
        \begin{tabular*}{\textwidth}{ccccccc}
            \topline
            Benchmark & $t_1$ & $t_2$ & $t_3$ & $t_4$ & $t_5$ & $t_6$ \midline
            \# prob. facts & 10 & 14 & 18 & 19 & 20 & 21 \\
            \# nodes circuit & 156 & 192 & 228 & 462 & 750 & 1465
            \botline
        \end{tabular*}
    \end{minipage}
    \caption{Inference time on benchmarks.}
    \label{fig:inference}
\end{figure}

We consider a variation of a typical PLP example where a set of people has a certain probability of having asthma or being stressed, and stress leads with some probability to smoking:
\[
    \begin{array}{l}
        0.1::asthma(X)\leftarrow \mathit{person}(X). \\
        0.3::stress(X)\leftarrow \mathit{person}(X). \\
        0.4::smokes(X)\leftarrow stress(X).
    \end{array}
\]
People are related by an influence relation: if a person smokes and influences to some extent another one, then the other person will smoke, and if someone smokes there is a probability to suffer from asthma. If someone suffers from asthma then the person does not smoke (an example of a cycle with negation).
\[
    \begin{array}{l}
        smokes(X) \leftarrow \mathit{influences}(Y,X), smokes(Y). \\
        0.4::asthma(X) \leftarrow smokes(X).                      \\
        \lnot smokes(X) \leftarrow asthma(X).
    \end{array}
\]
With this last rule we add a cycle through negation to a classical standard benchmark used in the Statistical Relational Learning community. This benchmark is often used to study the scaling capabilities of frameworks by growing the size of the domain of people.
Let $R$ be the aforementioned rules, we consider examples with an increasing number of people and relationships:
\begin{itemize}
    \item $t_1$ = $R \cup \{\mathit{person}(1). \;\;\mathit{person}(2). \;\; 0.3::\mathit{influences}(1,2).\;\; 0.6::\mathit{influences}(2,1).\}$
    \item $t_2=t_1 \cup \{\mathit{person}(3).\}$
    \item $t_3=t_2 \cup \{\mathit{person}(4).\}$
    \item $t_4=t_3 \cup \{0.2::\mathit{influences}(2,3).\}$
    \item $t_5=t_4 \cup \{0.7::\mathit{influences}(3,4).\}$
    \item $t_6=t_5 \cup \{0.9::\mathit{influences}(4,1).\}$
\end{itemize}
Figure~\ref{fig:inference} shows the running time of \smp{} on the different benchmarks for the queries $\mathit{query(smokes(X)).}$ and $\mathit{query(asthma(X)).}$ The cost of grounding (step 1) is negligible, hence we omit it from the plot. This experiment answers question Q1: as expected, the enumeration step dominates the running time of \smp{} and it is directly related to the size of the circuit. A larger circuit is likely to present a higher number of Cartesian products in the enumeration step, hence a higher computational cost is expected. This experiment confirms empirically the expected exponential cost of computing weights for the \smp{} semantics on a standard d-DNNF, in contrast with the polynomial time required for the traditional $\mathit{WMC}$ task. At the same time, it provides an example of a problem where the optimization of the evaluation step presented in Section~\ref{sec:inference} prevents the repetition of the cost of the enumeration step. The cost of the evaluation step is in fact consistently lower than the enumeration step. Finally, we remark that the approach based on knowledge compilation allows us to answer an increasing number of (ground) queries: 4 for $t_1$, 6 for $t_2$ and 8 for the remaining benchmarks. This affects the evaluation time, but the grounding, compilation and enumeration steps are executed once, regardless of the number of queries.


\paragraph{Q2. Learning.}

We answer question 2 by considering a dataset of 283 argument graphs~\citep{DBLP:conf/lrec/StedeAPAP16} and by deriving from each annotation of the dataset a bipolar argument graph. We attach to arguments and relations random probabilities to reflect an agent's belief in each, as we did for the running example, Example~\ref{ex:arglp_base}. The graphs contain from 4 to 11 nodes (average $5.3$) with an average of $1.4$ attacks and $3$ supports per graph. \\
Given this dataset of probabilistic argument graphs, we test learning as follows. For each argumentation graph modelled in a program $\mathcal{L}$ we obtain the model to be learned $\mathcal{L}^*$ by replacing each probability with a (randomly initialized) learnable parameter $t(\_)$, similarly to Example~\ref{ex:base_learn}. The programs $\mathcal{L}^*$ contain on average $10$ learnable parameters divided between probabilistic facts (arguments' biases) and probabilistic rules (relations' belief).
We use $\mathcal{L}$ to sample $n$ observations of the argumentation graph by means of the sampling tool from ProbLog2. Then, we use $\mathcal{L}^*$ to learn the original probabilities from the samples. We evaluate the accuracy of the learned parameter by considering the Mean Absolute Error (MAE) of the learned probabilities in $\mathcal{L}^*$ compared to the original ones in $\mathcal{L}$. We run the learning test with increasing size of the samples, i.e. $n\in\{50,100,150,200,250,300\}$, and with an upper bound of 100 EM iterations.

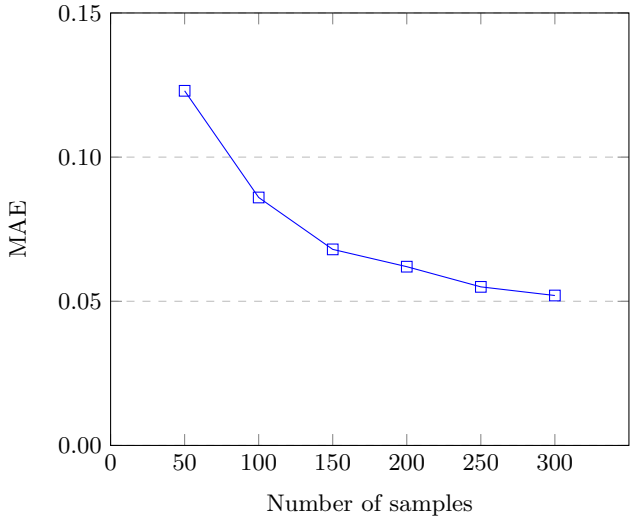
\begin{figure}[t]
    \begin{tikzpicture}
        \begin{axis}[
                y tick label style={
                        /pgf/number format/.cd,
                        fixed,
                        fixed zerofill,
                        precision=2,
                        /tikz/.cd
                    },
                xlabel={Number of samples},
                ylabel={MAE},
                xmin=0, xmax=350,
                ymin=0, ymax=0.15,
                xtick={0,50,100,150,200,250,300},
                ytick={0,0.05,0.1,0.15},
                legend pos=north west,
                ymajorgrids=true,
                grid style=dashed,
            ]

            \addplot[
                color=blue,
                mark=square,
            ]
            coordinates {
                    (50,0.123)(100,0.086)(150,0.068)(200,0.062)(250,0.055)(300,0.052)
                };

        \end{axis}
    \end{tikzpicture}
    \caption{Mean absolute error by number of samples.}
    \label{fig:learning_mae}

\end{figure}

Note that we are in the case of partial observability, as the parameters are attached to the predicates $bias$, $arg_{pos}$ and ${arg_{neg}}$, but we observe only the outcome in the form of the predicates $arg$.
We evaluate the quality of the learned programs with the mean absolute error (MAE), summarized in Figure~\ref{fig:learning_mae}.
From the experiments we observe that the MAE decreases with the increase of the sample size, that is, the more observations from the original distribution are provided, the higher is the accuracy in learning the original probabilities. At the same time when at least 100 samples are provided, the MAE drops below $10\%$. Therefore, we can answer positively to our question about learnability of beliefs in argument graphs with \smp{}.

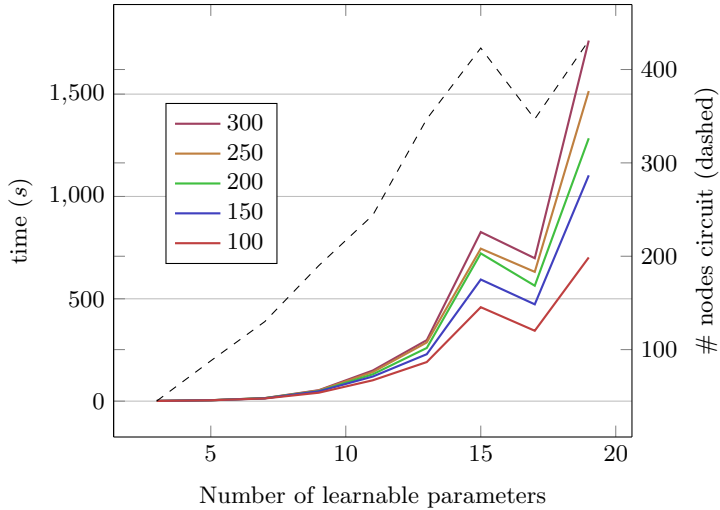
\begin{figure}[t]
    \centering
    \begin{tikzpicture}
        \begin{axis}
            [
                axis y line*=left,
                ymajorgrids=true,
                legend style={at={(0.1,0.4)},anchor=south west},
                legend cell align=left,
                ylabel={time ($s$)},
                xlabel={Number of learnable parameters},
            ]
            \addplot [thick,color=purple!50!gray]
            table[x=nparams, y=time300, col sep=tab]{learning_time.csv};
            \addplot [thick,color=orange!50!gray]
            table[x=nparams, y=time250, col sep=tab]{learning_time.csv};
            \addplot [thick,color=green!50!gray]
            table[x=nparams, y=time200, col sep=tab]{learning_time.csv};
            \addplot [thick,color=blue!50!gray]
            table[x=nparams, y=time150, col sep=tab]{learning_time.csv};
            \addplot [thick, color=red!50!gray]
            table[x=nparams, y=time100, col sep=tab]{learning_time.csv};
            \addlegendentry{300}
            \addlegendentry{250}
            \addlegendentry{200}
            \addlegendentry{150}
            \addlegendentry{100}

        \end{axis}
        \begin{axis}[
                ylabel near ticks, yticklabel pos=right,
                axis x line=none,
                ylabel={\# nodes circuit (dashed)},
                scaled y ticks = false
            ]
            \addplot [dashed]
            table[x=nparams, y=size, col sep=tab]{learning_time.csv};
        \end{axis}

    \end{tikzpicture}
    \caption{Mean running time by number of parameters and size of circuit (dashed line) on increasing number of observations (coloured lines).}
    \label{fig:learning_time}
\end{figure}

\paragraph{Q3. Learning vs. inference.}
Under the same experimental setup of \emph{Q2}, we now analyze the time required to learn the parameters with respect to the cost of inference, which is part of the EM learning algorithm (Equation~\ref{eq:learning}). There are two dimensions that influence the time required for learning with respect to inference, which we represent in Figure~\ref{fig:learning_time}.
The first is the cost of updating the probabilities for an increasing number of observations. In Figure~\ref{fig:learning_time} we show the running time (left scale) of EM learning on increasing sample sizes in ${100,150,200,250,300}$.
The second is the intrinsic cost of inference with respect to the complexity of the program. We measure the complexity of the program by measuring the size of the compiled circuit, represented in Figure~\ref{fig:learning_time} with a dashed line (right scale).

The experiment shows that the second dimension is the main factor in determining the time required for learning, which follows the same trend of inference shown in Figure~\ref{fig:inference}. Since we replace all probabilistic facts with learnable parameters, the x-axis of the two figures are comparable. The dashed line represents the average size of the circuit for a set of programs with a given number of learnable parameters (x-axis). As expected from Equation~\ref{eq:learning}, the experiment shows a strong correlation between the inference cost on an increasingly large circuit and the learning time. In particular, we observe that the average learning time with 17 parameters decreases from 15, because in the dataset the average size of the circuits with 17 parameters is also smaller.
The first dimension of the experiment shows that on smaller models the difference between the sample sizes (coloured lines) is negligible, while on more complex models the differences spread out. This is plausible because on more complex models the cost of inference is higher and thus the repetition of a larger cost over more examples becomes more evident. However, the impact of a higher number of observation on the total time required for learning is clearly inferior to a bigger (more complex) logical circuit.

\section{Related Work}
In this section we discuss the related work with respect to probabilistic logic programming and probabilistic argumentation.
\subsection{Probabilistic Logic Programming.}\label{sec:related_prob}
Both three-valued well-founded model semantics and stable model semantics have been considered to extend traditional PLP framework's semantics: \citeN{DBLP:conf/ismvl/HadjichristodoulouW12} present an extension of PRISM based on three-valued well-founded models.
When considering general normal logic programs, well-founded semantics~\citep{DBLP:journals/jacm/GelderRS91} is a three-valued semantics, that is, logical atoms can be true, false or undefined. A three-valued interpretation is a pair $I=(T,F)$ where $T$ and $F$ are disjoint subsets of the Herbrand base $HB(\mathcal{L})$ of the program $\mathcal{L}$. The (ground) atoms in $T$ (resp. $F$) are interpreted as \emph{true} (resp. \emph{false}) in $I$. The remaining atoms in $HB(\mathcal{L})\backslash(T\cup F)$ are said to be \emph{undefined}. A three-valued model is thus a three-valued interpretation which satisfies all the rules of the program. Our approach is different because with stable model semantics we assign a truth value to the non-deterministic choices that well-founded semantics would label as undefined. At the same time we use the third value to denote the absence of stable models.

On the other hand, the application of stable model semantics in PLP is represented by probabilistic answer set programming (ASP) \citep{DBLP:journals/jair/CozmanM17,DBLP:journals/ijar/CozmanM20}. Table~\ref{tab:compareplp} summarizes the differences between the frameworks, which emerge on the definition of the probability distribution and the definition of model of a possible world.
\begin{table}[t]
    \caption{PLP frameworks comparison. n.a. = not applicable. $2^*$= the logic semantics is two-valued stable models, but we introduce a third value for total choices with 0 stable models.}
    \label{tab:compareplp}
    \centering
    \begin{tabular}{ c  p{1.6cm}  c  p{1cm}  p{1.5cm} }
        \topline
        Framework                                           & Distribution semantics       & Model semantics               & Logic values             & Models per total choice  \midline
        ProbLog2~\citep{DBLP:conf/uai/FierensBTGR11}        & \hfil\cmark                  & Well-founded                  & \hfil 2                  & \hfil 1                           \\
        WF-PRISM                                            & \hfil\multirow{2}{*}{\cmark} & \multirow{2}{*}{Well-founded} & \hfil \multirow{2}{*}{3} & \hfil \multirow{2}{*}{1}          \\
        \citep{DBLP:conf/ismvl/HadjichristodoulouW12}       &                              &                               &                                                              \\
        P-Log~\citep{DBLP:journals/tplp/BaralGR09}          & \hfil\xmark                  & Stable models                 & \hfil 2                  & \hfil n.a.                        \\
        LP\textsuperscript{MLN}~\citep{DBLP:conf/kr/LeeW16} & \hfil\xmark                  & Stable models                 & \hfil 2                  & \hfil n.a.                        \\
        \smp{}                                              & \hfil \cmark                 & Stable models                 & \hfil $2^*$              & $\hfil \geq 1$
        \botline
    \end{tabular}
\end{table}

\paragraph{Relation with probabilistic ASP.}
The key difference between \smp{} and probabilistic ASP frameworks is that our semantics is based on the distribution semantics, that is, we define a probability distribution over models from a probability distribution over total choices.
On the contrary, probabilistic ASP languages such as P-Log~\citep{DBLP:journals/tplp/BaralGR09} and LP\textsuperscript{MLN}~\citeN{DBLP:conf/kr/LeeW16}  use the probability labels to directly define a (globally normalized) probability distribution over the models of a (derived) program. This difference allows us to preserve the marginal probability of independent facts in the joint distribution in the presence of multiple stable models, and to measure the probability of an inconsistency in a program. In fact, following the probabilistic ASP approach can give counterintuitive results as we show in Example~\ref{ex:global}.
\begin{example}\label{ex:global}
    In Example~\ref{ex:multiple}, if we associate to each stable model $M_{\omega}$ the probability of the corresponding total choice $P(\omega)$, the sum of the probabilities of the stable models is $5\cdot0.25=1.25$. A normalization w.r.t. all answer sets, as P-Log does, leads to a probability distribution over stable models $\hat{P}'(M_{\omega})=0.25/1.25=0.2$ for all $M_{\omega}$. Then the probability of a query $\bg$ is $\mathbb{P}(\bg)=\hat{P}'(M_{\omega_1})+\hat{P}'(M_{\omega_2})=0.4$. This means that the marginal probability of $\bg$ in the joint distribution is lower than the prior 0.5 despite no epistemic influence is defined on $\bg$. In our approach, the  marginal probability of that fact being true under the final distribution equals the fact's label, and is thus directly interpretable.
    Similarly, when some possible worlds are inconsistent, the corresponding probability mass is not extended to a model, but rather a global normalization redistributes uniformly such mass to the valid models. For this reason, in Example~\ref{ex:zero} a global normalization would assign all the probability mass to $\omega_4$, i.e. $\hat{P}'(\omega_4) = \frac{0.25}{0.25}=1$. This because $\{\mathit{right}\}$ is the only possible stable model hence its weight, $0.25$, is also the sum of all possible stable models.
    Therefore, all atoms but $\mathit{right}$ would be interpreted false and $\mathit{right}$ would be interpreted true with probability 1, thus making the presence of logical inconsistencies unclear.
\end{example}

\begin{example}\label{ex:global_minus}
    In Example~\ref{ex:hotels_sm} we have a similar situation: by weighing each stable model with the probability labels of the literals in the model we have that the sum of the weights of the models is $1.42$ and that $\hat{P}'(M_{\omega_1})=\frac{0.12}{1.42}=0.085$,
    $\hat{P}'(M_{\omega_2})=\frac{0.18}{1.42}=0.1268$, $\hat{P}'(M_{\omega_3})=\frac{0.28}{1.42}=0.1972$, $\hat{P}'(M^1_{\omega_4})=\hat{P}'(M^2_{\omega_4})=\frac{0.42}{1.42}=0,2958$. Therefore, the probabilities of $\te$ and $\tn$ are respectively $\hat{P}'(M_{\omega_2})+\hat{P}'(M^1_{\omega_4})+\hat{P}'(M^2_{\omega_4})=0.7183$ and $\hat{P}'(M_{\omega_3})+\hat{P}'(M^1_{\omega_4})+\hat{P}'(M^2_{\omega_4})=0.7887$. Therefore, the probability of probabilistic facts in the inferred distribution increases w.r.t. the original declared value. This happens with a globally normalized definition of the probability of the models and multiple stable models per total choice.
\end{example}
In this paper we focus on computing exact probabilities as \citeN{DBLP:journals/tplp/BaralGR09} and~\citeN{DBLP:conf/kr/LeeW16} rather than defining an interval of probabilities for each atom as in \emph{credal semantics}~\citep{DBLP:conf/ecai/Lukasiewicz98}.
\citeN{DBLP:journals/ijar/CozmanM20} remarks the difference between a credal semantics approach and a distribution semantics approach like ours. They emphasize that in distribution semantics a probabilistic fact $p::f.$ the fact $f$ is imposed with probability $p$ and discarded with probability $1-p$, while their approach distinguishes between taking $f$ with probability $p$ and $\lnot f$ with probability $1-p$. Moreover, credal semantics are defined only for consistent logic programs, as in~\cite{DBLP:conf/lpnmr/AzzoliniBR22}. Recent work~\citep{DBLP:conf/kr/RochaC22} extends credal semantics to L-Credal semantics to handle inconsistencies. We illustrate the differences by means of the graph colouring example from \citeN{DBLP:conf/ilp/CozmanM16}, other comparisons are also present in~\citeN{DBLP:conf/kr/RochaC22}.

\begin{example}\label{ex:credal}
    Consider the following program:\vspace{-1em}
    \begin{program}
        \begin{array}{l l l l l}
            \multicolumn{5}{c}{coloredBy(V, red) \leftarrow {\sim} coloredBy(V, yellow), {\sim} coloredBy(V, green), vertex(V ).} \\
            \multicolumn{5}{c}{coloredBy(V, yellow) \leftarrow {\sim} coloredBy(V, red), {\sim} coloredBy(V, green), vertex(V ).} \\
            \multicolumn{5}{c}{coloredBy(V, green) \leftarrow {\sim} coloredBy(V, red), {\sim} coloredBy(V, yellow), vertex(V ).} \\
            \multicolumn{5}{c}{noClash \leftarrow {\sim} noClash, edge(V, U ), coloredBy(V, C), coloredBy(U, C).}                 \\
            edge(1, 4). & edge(2, 1).      & edge(2, 4).        & edge(3, 5).          & edge(4, 3).                              \\
            edge(1, 3). & 0.5::edge(4, 5). & coloredBy(2, red). & coloredBy(5, green). &                                          \\
        \end{array}
    \end{program}
    Credal semantics here define probability intervals based on the stable models of the two cases where $edge(4,5)$ is present (one model) or not (two models). The possible colorings are represented in Figure~\ref{fig:credal}.  Credal semantics define a lower ($\underline{\mathbb{P}}$) and upper ($\overline{\mathbb{P}}$) probability for each node: $\underline{\mathbb{P}}(\mathit{coloredBy}(1,\mathit{yellow})) = 0$, $\overline{\mathbb{P}}(\mathit{coloredBy}(1,\mathit{yellow})) = \frac{1}{2}$, $\underline{\mathbb{P}}(\mathit{coloredBy}(4,\mathit{yellow})) = \frac{1}{2}$, $\overline{\mathbb{P}}(\mathit{coloredBy}(4,\mathit{yellow})) = 1$, and $\underline{\mathbb{P}}(\mathit{coloredBy}(3,\mathit{red})) = \overline{\mathbb{P}}(\mathit{coloredBy}(3,\mathit{red})) = 1$.
    On the contrary with our semantics we compute exact values: $\mathbb{P}(\mathit{coloredBy}(1,\mathit{yellow}))=0.25$, $\mathbb{P}(\mathit{coloredBy}(4,\mathit{yellow}))=0.75$, $\mathbb{P}(\mathit{coloredBy}(3,\mathit{red}))=1$. These values derive from the probability of $0.5$ of the coloring with $edge(4,5)$ and the probability of $\frac{0.5}{2}$ of each of the two possible coloring without $edge(4,5)$.
\end{example}

\begin{figure}[t]
    \begin{subfigure}[t]{0.24\textwidth}
        \centering
        \begin{tikzpicture}[->]
            \tikzstyle{every state}=[fill=white,shape=circle,minimum size=0.5cm]

            \node[state] (A) {1};
            \node[state]  (B) [below=0.5cm of A, fill=red] {2};
            \node[state]  (C) [right=0.5cm of A] {3};
            \node[state]  (D) [below=0.5cm of C]  {4};
            \node[state]  (E) [right=0.5cm of C,fill=green]  {5};

            \path[->] (A) edge []  node {} (C);
            \path[->] (A) edge []  node {} (D);
            \path[->] (B) edge []  node {} (A);
            \path[->] (B) edge []  node {} (D);
            \path[->] (C) edge []  node {} (E);
            \path[->] (D) edge []  node {} (C);
            \path[->, dashed] (D) edge []  node {} (E);
        \end{tikzpicture}
        \caption{Graph with $0.5::edge(4,5)$}
        \label{fig:coloring_graph}
    \end{subfigure}
    \begin{subfigure}[t]{0.24\textwidth}
        \centering
        \begin{tikzpicture}[]
            \tikzstyle{every state}=[fill=white,shape=circle,minimum size=0.5cm]

            \node[state]  (A) [fill=green] {1};
            \node[state]  (B) [below=0.5cm of A, fill=red] {2};
            \node[state]  (C) [right=0.5cm of A, fill=red] {3};
            \node[state]  (D) [below=0.5cm of C, fill=yellow]  {4};
            \node[state]  (E) [right=0.5cm of C,fill=green]  {5};

            \path[->] (A) edge []  node {} (C);
            \path[->] (A) edge []  node {} (D);
            \path[->] (B) edge []  node {} (A);
            \path[->] (B) edge []  node {} (D);
            \path[->] (C) edge []  node {} (E);
            \path[->] (D) edge []  node {} (C);
            \path[->] (D) edge []  node {} (E);
        \end{tikzpicture}
        \caption{One coloring with $edge(4,5)$.}
        \label{fig:coloring_graph}
    \end{subfigure}
    \begin{subfigure}[t]{0.24\textwidth}
        \centering
        \begin{tikzpicture}[]
            \tikzstyle{every state}=[fill=white,shape=circle,minimum size=0.5cm]

            \node[state]  (A) [fill=green] {1};
            \node[state]  (B) [below=0.5cm of A, fill=red] {2};
            \node[state]  (C) [right=0.5cm of A, fill=red] {3};
            \node[state]  (D) [below=0.5cm of C, fill=yellow]  {4};
            \node[state]  (E) [right=0.5cm of C,fill=green]  {5};

            \path[->] (A) edge []  node {} (C);
            \path[->] (A) edge []  node {} (D);
            \path[->] (B) edge []  node {} (A);
            \path[->] (B) edge []  node {} (D);
            \path[->] (C) edge []  node {} (E);
            \path[->] (D) edge []  node {} (C);
        \end{tikzpicture}
        \caption{First coloring without $edge(4,5)$.}
        \label{fig:coloring_graph}
    \end{subfigure}
    \begin{subfigure}[t]{0.24\textwidth}
        \centering
        \begin{tikzpicture}[]
            \tikzstyle{every state}=[fill=white,shape=circle,minimum size=0.5cm]

            \node[state]  (A) [fill=yellow]{1};
            \node[state]  (B) [below=0.5cm of A, fill=red] {2};
            \node[state]  (C) [right=0.5cm of A, fill=red] {3};
            \node[state]  (D) [below=0.5cm of C, fill=green]  {4};
            \node[state]  (E) [right=0.5cm of C,fill=green]  {5};

            \path[->] (A) edge []  node {} (C);
            \path[->] (A) edge []  node {} (D);
            \path[->] (B) edge []  node {} (A);
            \path[->] (B) edge []  node {} (D);
            \path[->] (C) edge []  node {} (E);
            \path[->] (D) edge []  node {} (C);
        \end{tikzpicture}
        \caption{Second coloring without $edge(4,5)$.}
        \label{fig:coloring_graph}
    \end{subfigure}
    \caption{Possible colorings of Example~\ref{ex:credal}}.
    \label{fig:credal}
\end{figure}

\paragraph{Relations with traditional PLP.} Our approach generalizes traditional PLP frameworks on programs without function symbols: when a program defines total choices corresponding to exactly one two-valued well-founded model, then the semantics agree on both models and probability.
As for the model, if a normal logic program has a total two-valued well-founded model, then the model is the unique stable model~\citep{DBLP:journals/jacm/GelderRS91}. Thus, in this case $|M(\omega)|=1$ for each total choice $\omega$, from which it follows that $\hat{P}(M_{\omega})=P(\omega)$ for the single stable model $M_{\omega}\in \mathit{MOD}(\mathcal{L},\omega)$ ($\mathit{MOD}(\mathcal{L},\omega) = \{M_{\omega}\}$). This means that the probability of the model is the probability of the corresponding total choice as in (probabilistic) two-valued well-founded semantics. The one-to-one correspondence between total choices and stable models also guarantees that $\mathbb{P}((\emptyset,\emptyset))=0$.

For this reason, our pipeline is also a generalization of ProbLog2's pipeline.
ProbLog2~\citep{DBLP:conf/uai/FierensBTGR11} reduces the probabilistic inference task to a \emph{weighted model counting problem}~\citep{DBLP:journals/aicom/CadoliD97} in three steps:
\begin{enumerate}
    \item Compute the relevant grounding, that is, ground only the part of the program necessary to answer the query.
    \item Convert the ground rules into an equivalent boolean formula (CNF).
    \item Compile the boolean formula into an arithmetic circuit to efficiently compute the weighted model count of the formula.
\end{enumerate}
The grounding procedure in ProbLog2 computes the \emph{relevant} ground program w.r.t. the given queries $Q$ and evidence $E$. The relevant ground program is obtained by applying SLD resolution to prove all atoms in $Q\cup E$. However, in \smp{} an atom can be derived from a logical choice whose atoms are not directly justified by some (probabilistic) fact, therefore in this case meaningful parts of the program would not be included in the relevant ground program. This is why we use a standard bottom-up technique.
\citeN{DBLP:conf/aaai/AzizCMS15} apply stable model counting techniques in ProbLog (\textsc{asp}ProbLog). In particular, they replace the \textsc{Dsharp} compiler~\citep{DBLP:conf/ai/MuiseMBH12} used to implement step 3 with a stable model knowledge compiler (an extension of \textsc{Dsharp} itself). Step 2 is also modified, since the CNF conversion is skipped and a simple representation of the ground rules is provided to the stable model knowledge compiler. However, \textsc{asp}ProbLog does not change the semantics of ProbLog, and invalid programs according to the original semantics are still rejected. Moreover, \textsc{asp}ProbLog does not implement the syntactical features of ProbLog2 on which our approach to argumentation is based, such as negation in the heads or annotated rules.

Distribution semantics has been proven by~\citeN{DBLP:conf/iclp/Sato95} to be well-defined for definite programs with function symbols, and later~\citeN{DBLP:conf/iclp/Riguzzi15} proves this for the case of normal programs with two-valued well-founded models. Studying the semantics of \smp{} programs with function symbols is thus an interesting direction for future work.

\subsection{Probabilistic argumentation.}

\begin{table}[b]
    \caption{Argumentation frameworks comparison. $\sim{}$ denotes partial support}
    \label{tab:compare}
    \centering
    \begin{tabular}{ c  c  c   p{2cm}  p{1.6cm} }
        \topline
        Framework                                            & Epistemic               & PLP                     & Conditionals \& Marginals    & \hspace{0pt}Implementation \midline
        QBAF~\citep{DBLP:journals/ijar/BaroniRT19}           & \xmark                  & \xmark                  & \hfil\xmark                  & \hfil\xmark                         \\
        Prob. argument graphs~\citep{DBLP:conf/tafa/LiON11}  & \xmark                  & \xmark                  & \hfil\xmark                  & \hfil\cmark                         \\
        MetaProblog                                          & \multirow{2}{*}{\xmark} & \multirow{2}{*}{\cmark} & \hfil\multirow{2}{*}{\xmark} & \hfil\multirow{2}{*}{\cmark}        \\
        \citep{DBLP:journals/ijar/MantadelisB20}             &                         &                         &                              &                                     \\
        Epistemic graphs~\citep{DBLP:journals/ai/HunterPT20} & \cmark                  & \xmark                  & \hfil\xmark                  & \hfil\cmark                         \\
        L-Credal~\citep{DBLP:conf/kr/RochaC22}               & \cmark                  & \cmark                  & \hfil $\sim{}$               & \hfil \xmark                        \\
        \smp{}                                               & \cmark                  & \cmark                  & \hfil\cmark                  & \hfil\cmark
        \botline
    \end{tabular}
\end{table}

Our approach is novel as previous work considers different semantics and reasoning techniques for probabilistic argument graphs.  Table~\ref{tab:compare} summarizes these differences: the choice between interpreting probabilities as in the constellation approach or as in the epistemic approach, approaching the problem from PLP modelling a single joint probability distribution as in Bayesian networks, whether it can reason about marginal and conditional queries about the probability of arguments, and the implementation in a framework where the probabilities of the graph can be learned. Moreover, our framework is the only one providing an implementation that can learn from observations (of accepted arguments).

\paragraph{Epistemic.} We follow the epistemic interpretation of probabilities~\citep{DBLP:journals/ai/HunterPT20,DBLP:journals/ijar/Hunter13} as a direct measure of the belief of an agent in arguments. Section~\ref{sec:joint} already discussed the main differences with previous work. We use fine-grained (gradual) evaluations of relations and base scores for arguments in a bipolar setting similarly to Quantified Bipolar Argument Frameworks (QBAFs)~\citep{DBLP:journals/ijar/BaroniRT19}.
Similarly to our framework, in a QBAF a set of arguments contains both an attack and support relation, and defines a base score $\tau$, which corresponds in our case to the prior probability (belief) $P_A$, as \citeN{DBLP:journals/ijar/BaroniRT19} describes it: ``\dots the nature and meaning
of the base score $[\ldots]$ corresponds to an assessment of arguments which precedes the consideration of the relations of attack and support with other arguments''. In our case the joint probability distribution is the result of this consideration.
QBAF is a restricted setting compared to our framework since we also define a score (following the terminology of~\citeN{DBLP:journals/ijar/BaroniRT19}) for the relations $R^+$ and $R^-$, besides considering set-attacks. Most importantly, we do this with probabilistic semantics.

\paragraph{PLP.} At the same time, we propose a mapping from probabilistic argument graphs to probabilistic logic programs, which provides a novel semantics for probabilistic argument graphs. We consider a single probability distribution in a Bayesian style rather than families of distributions.
Previous work in fact focused on reasoning about the properties of families of probability distributions that are consistent with the argument graph and additional constraints (\emph{epistemic graphs})~\citep{DBLP:journals/ai/HunterPT20}.

Despite following the epistemic approach, our framework has similarities with the constellations approach. The definition of the distribution over subgraphs in the constellations approach is similar to the distribution over subprograms $F\cup R$ of distribution semantics. In fact, the distribution modelled by the probabilistic logic program is determined by the probability that an argument is included in a possible world, similar to the inclusion or exclusion of nodes and edges in the subgraphs considered in the constellations approach. However, in order to determine whether an argument is true or not, we rely on the stable model semantics for logic programs.
In the constellations approach, on the other hand, the admissible arguments are defined by the classical extension-based semantics~\citep{DBLP:journals/ai/Dung95}.
Stable models are equivalent to stable extensions for basic arguments graphs~\citep{DBLP:journals/ai/Dung95}.
We consider however more sophisticated relations between arguments that cannot be encoded in a traditional (probabilistic) abstract argumentation framework, e.g. support~\citep{DBLP:conf/nmr/AmgoudCL04} or set-attacks~\citep{DBLP:conf/argmas/NielsenP06}. For this reason, our approach does not fit any of the previous categorizations about the connections between logic (programming) and argumentation~\citep{DBLP:journals/argcom/BesnardCL20}.

Modelling (deterministic) argumentation problems by means of logic programming dates back to the foundational work of~\citeN{DBLP:journals/ai/Dung95}.~\citeN{DBLP:journals/ai/Dung95} proposes a method to define meta-interpreters for argument systems encoding the extension-based semantics as logic rules. A change in argumentation semantics thus requires to encode in the meta-interpreter a new reasoning technique. This is the approach taken in~\citeN{DBLP:journals/ijar/MantadelisB20}, while we directly compute the stable models semantics for logic programs on a mapping from argument graphs to probabilistic logic programs.

\citeN{DBLP:journals/sLogica/WuCG09} propose a mapping for Dung's abstract argumentation frameworks to logic programs, therefore they do not address the probabilistic settings as well as other extensions to the original framework considered in this paper. \citeN{DBLP:journals/sLogica/WuCG09} focus on complete labellings and propose a translation where given an argument graph $(A,R)$, given a set of attacks of the form $(b_i,a)\in (A,R)$, $i\in\{1,\dots,n\}$ a logic rule of the form $a\leftarrow {\sim}b_1, \dots, {\sim}b_n.$ is added to the logic program translation. This is a similar principle followed in our encoding, where attacks $(b_i,a)$ with $P_R((b_i,a))=p_i$ are translated to rules of the form $p_i::\lnot a\leftarrow b_i.$ In fact, the rewriting of negation in the head produces rules $a \leftarrow a_{pos}, {\sim}a_{not}$ plus $p_i::a_{not} \leftarrow b_i$. Therefore, in both cases all attackers $b_i$ must be false in order to consider $a$ acceptable. However, there is a difference between negation in the head (and the corresponding translation) and negation in the body. Attacks directly encoded with negation in the body do not exclude that the two choices can be true at the same time because of external justification, but exclude that both are false. Vice versa, an encoding with negation in the head excludes that both choices are true at the same time, but not that both can be false (e.g. in Example~\ref{ex:hotel_sm_math}).\newpage

\begin{example}\label{ex:negs_pws}
    Consider the difference between a program with a loop with negation in the body (A) and one with negation in the head (B).
    \begin{program}
        \begin{array}{ l | l}
            \textbf{A}               & \textbf{B}          \\
            0.6::\te.                & 0.6::\te.           \\
            0.7::\tn.                & 0.7::\tn.           \\
            \hy \leftarrow \te.      & \hy \leftarrow \te. \\
            \hx \leftarrow \tn.      & \hx \leftarrow \tn. \\
            \lnot\hx \leftarrow \hx. &
            \hx \leftarrow {\sim}\hy.                      \\
            \lnot\hy \leftarrow \hx. &
            \hy \leftarrow {\sim}\hx.
        \end{array}
    \end{program}
    Program A is the program of Example~\ref{ex:hotel_sm_math}. Program B is a similar encoding with negation in the body. In this case the possible worlds $\omega_2$ and $\omega_3$ have the same models of A. However, for $\omega_1=\{\}$ where program A has an empty model, program B has the models $\{\{\hy\}, \{\hx\}\}$. While program B forces a choice when the counterarguments are not accepted, program A allows no choice. At the same time, program B allows on $\omega_4$ to infer a stay at both hotels from the counterarguments for each, that is, the model $\{\te, \tn, \hx, \hy\}$. Table~\ref{tab:negs_pws} compares the models for each total choice.
    \begin{table}[b]
        \caption{Possible models comparison of Example~\ref{ex:negs_pws}}
        \label{tab:negs_pws}
        \centering
        \begin{tabular}{ l c c}
            \topline
            Total choice                             & Model A                                  & Model B                                \\\hline
            $\omega_1=\{\}$                          & $\{\}$                                   & $\{\{\hy\}, \{\hx\}\}$                 \\
            $\omega_2=\{\te\}$                       & $\{\omega_2\cup\{\hy\}\}$                & $\{\omega_2\cup\{\hy\}\}$              \\
            $\omega_3=\{\tn\}$                       & $\{\omega_3\cup\{\hx\}\}$                & $\{\omega_3\cup\{\hx\}\}$              \\
            \parbox{2cm}{\begin{align*}
                    \omega_4 = \{ & \te,  \\
                                  & \tn\}
                \end{align*}} & \parbox{2cm}{\begin{align*}
                    \{ & \omega_4\cup\{\hy\},  \\
                       & \omega_4\cup\{\hx\}\}
                \end{align*}} & $\{\omega_4\cup\{\hy, \hx\}\}$\botline \\
        \end{tabular}
    \end{table}
\end{example}

Past work on the connections with logic programming beyond basic abstract argumentation frameworks are limited to the deterministic setting,
except for the work by \cite{DBLP:conf/kr/RochaC22}. They adopt the translation by \cite{DBLP:journals/sLogica/WuCG09} to encode Assumption Based Argumentation Frameworks~\citep{DBLP:conf/lpnmr/BondarenkoTK93} in a probabilistic logic program.
\citeN{DBLP:journals/ijar/CaminadaSAD15} explore equivalence relations for more expressive deterministic argumentation frameworks, by framing logic programming semantics in terms of argumentation semantics.
\citeN{DBLP:journals/tplp/AlfanoGPT20} consider the opposite approach and follow in a \emph{deterministic} setting a similar approach to ours for \emph{probabilistic} argumentation. In fact, a simple but general logical framework is shown to be able to capture, in a systematic and succinct way, the different features of several argumentation frameworks under different argumentation semantics. The authors remark how the flexibility of a logic programming approach encourages the study of more extensions and can be used for better understanding the semantics of extended argumentation frameworks.

\paragraph{Marginals and conditionals.} With our methodology it is possible to apply general PLP reasoning techniques to probabilistic argument graphs, such as marginal probability computation (Example~\ref{ex:marg_arg}), conditioning over evidence (Example~\ref{ex:evidence_arg}), and parameter learning (Section~\ref{sec:learning}).
This results in a modular, expressive, extensible framework reflecting the dynamic nature of beliefs in an argumentation process. On the contrary, argumentation systems like epistemic graphs require (ad-hoc) reasoning algorithms that do not follow the traditional probabilistic reasoning techniques. In the framework by \cite{DBLP:conf/kr/RochaC22} conditioning on atoms that appear negated in a body is deferred to future work.

\paragraph{Implementation.} In Section~\ref{sec:implementation} we presented a system that computes \smp{} semantics, while in the past the focus has been on developing new argumentation semantics rather than tools to compute them. Moreover, we provide an implementation of the EM learning algorithm for probabilistic parameters whose application to probabilistic argumentation problems is novel.

\section{Conclusion}

We proposed a novel approach to epistemic probabilistic argumentation in two points. First, the interpretation of a probabilistic argument graph as a probabilistic graphical model defining a joint probability distribution over arguments. Second, we presented a novel syntactical translation of argument attacks to logic programming rules based on negation in the head and the inhibition effect.

Our method regards probabilistic argument graphs as the description of how the prior beliefs (biases) interact with each other to define marginal and conditional beliefs that are coherent with the argumentative (logic) structure. This allows us to exploit the structure of conditional (in)dependencies of arguments and apply the traditional PLP inference methods to the argumentative representation of subjective beliefs, such as learning, conditional queries, and most probable explanations.

Approaching probabilistic argumentation from a probabilistic logic programming perspective stresses the limiting assumptions of PLP frameworks when (probabilistic) normal logic programs are concerned. For this reason in this paper we proposed a new PLP system, \smp{}, based on a combination of the classical distribution semantics for probabilistic logic programs with stable model semantics.

\smp{} generalizes previous work on distribution semantics for programs without function symbols, supporting inference and parameter learning tasks for a wider class of (probabilistic) logic programs.
The experiments in this regard show that the existing PLP inference techniques can be used, but under the new semantics not all the necessary information can be computed efficiently. Novel knowledge compilation techniques allowing polynomial-time querying under \smp{}'s semantics would provide a significant improvement for both the inference and learning tasks, therefore future research on this topic represents an interesting and useful development of this work.


\section*{Acknowledgements}
This work was supported by the FWO project N. G066818N and the Flanders AI program.

\section*{Competing interests}
The authors declare none.

\bibliographystyle{acmtrans}
\bibliography{biblio}

\end{document}